\documentclass[sigconf,xcolor,dvipsnames,usenames]{acmart}

\usepackage{enumitem}
\usepackage{cleveref}
\usepackage{fancyvrb}
\usepackage{multirow}
\usepackage{pstricks}
\usepackage{algorithm}
\usepackage[noend]{algpseudocode}
\algrenewcommand\alglinenumber[1]{#1}
\algrenewcommand\algorithmicindent{1.0em}%
\usepackage{subcaption}
\usepackage[htt]{hyphenat}

\AtBeginDocument{%
  \providecommand\BibTeX{{%
    \normalfont B\kern-0.5em{\scshape i\kern-0.25em b}\kern-0.8em\TeX}}}

\setcopyright{none}
\copyrightyear{}
\acmYear{}
\acmDOI{}
\renewcommand\footnotetextcopyrightpermission[1]{}
\newcommand{\header}[1]{\vspace{1mm}\noindent\textbf{#1}.}
\newcommand{\headerl}[1]{\vspace{1mm}\noindent\textit{#1}.}
\newcommand{\headeru}[1]{\vspace{1mm}\noindent\underline{#1}.}

\usepackage{tikz}

\newcommand{\method}{\textsc{AnyMatch}}
\begin{document}

\title{\method{} -- Efficient Zero-Shot Entity Matching with~a~Small~Language~Model}

\author{Zeyu Zhang}
\affiliation{%
  \institution{University of Amsterdam \& Amsterdam UMC}
  \country{}
}
\email{z.zhang2@uva.nl}

\author{Paul Groth}
\affiliation{%
  \institution{University of Amsterdam}
  \country{}
}
\email{p.t.groth@uva.nl}

\author{Iacer Calixto}
\affiliation{%
  \institution{Amsterdam UMC, University of Amsterdam}
  \country{}
}
\email{i.coimbra@amsterdamumc.nl}

\author{Sebastian Schelter}
\affiliation{%
  \institution{BIFOLD \& TU Berlin}
  \country{}
}
\email{schelter@tu-berlin.de}

\begin{abstract}
Entity matching (EM) is the problem of determining whether two records refer to same real-world entity, which is crucial in data integration, e.g., for product catalogs or address databases. A major drawback of many EM approaches is their dependence on labelled examples. We thus focus on the challenging setting of \textit{zero-shot entity matching} where no labelled examples are available for an \textit{unseen target dataset}. Recently, large language models (LLMs) have shown promising results for zero-shot EM, but their low throughput and high deployment cost limit their applicability and scalability.

We revisit the zero-shot EM problem with \method{}, a small language model fine-tuned in a transfer learning setup. We propose several novel data selection techniques to generate fine-tuning data for our model, e.g., by selecting difficult pairs to match via an AutoML filter, by generating additional attribute-level examples, and by controlling label imbalance in the data.

We conduct an extensive evaluation of the prediction quality and deployment cost of our model, in a comparison to thirteen baselines on nine benchmark datasets. We find that \method{} provides competitive prediction quality despite its small parameter size: it achieves the second-highest F1 score overall, and outperforms several other approaches that employ models with hundreds of billions of parameters. Furthermore, our approach exhibits major cost benefits: the average prediction quality of \method{} is within 4.4\% of the state-of-the-art method \texttt{MatchGPT} with the proprietary trillion-parameter model \texttt{GPT-4}, yet \method{} requires four orders of magnitude less parameters and incurs a 3,899 times lower inference cost (in dollars per 1,000 tokens).
\end{abstract}

\maketitle
\pagestyle{plain}


\section{Introduction}
Entity matching (EM), often also referred to as deduplication or entity resolution, is the problem of determining whether two records refer to the same real-world entity. EM is a well-studied problem~\cite{li2020deep,doan2020magellan,chen2021gnem,fu2021hierarchical,papadakis2023critical,wang2022machop,mudgal2018deep} and of high practical importance in data integration~\cite{abedjan2016detecting,stonebraker2018data,gao2018navigating,zhang2020finding,huang2024relationalizing}.

\header{Zero-shot entity matching} A typical restriction in entity matching scenarios is the dependence on labelled examples. A less restrictive yet more challenging setting is {\em zero-shot entity matching}~\cite{wu2020zeroer}, where the matcher has to work with an unseen target dataset for which no labelled examples are available.
\begin{figure}[t!]
  \centering
  \includegraphics[width=0.9\columnwidth]{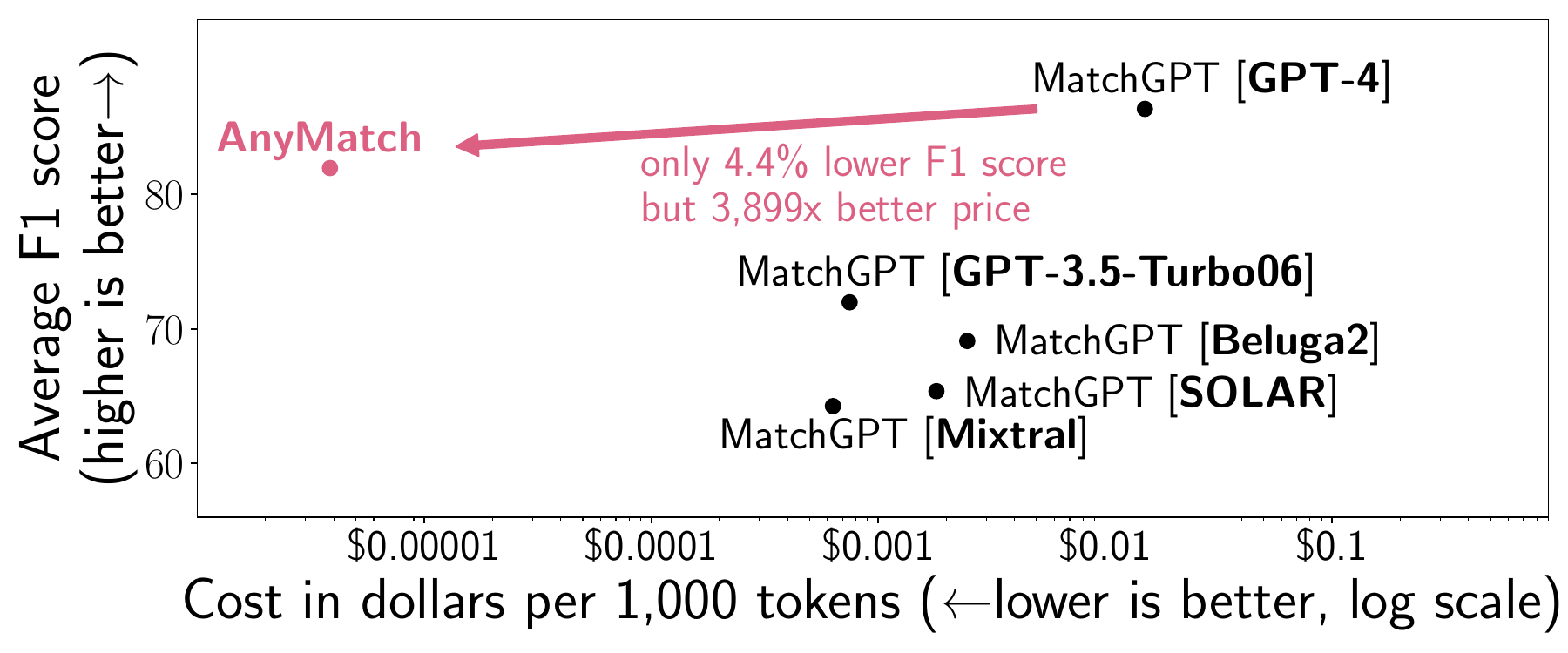}
  \caption{\method{} offers competitive zero-shot entity matching performance at a low cost. Its average F1 score is only 4.4\% lower than the score of the state-of-the-art method \texttt{MatchGPT} with OpenAI's \texttt{GPT-4} model, which requires a trillion parameter LLM at a 3,899x higher cost per 1,000 tokens.}
  \label{fig:teaser}
\end{figure}
Zero-shot matchers are essential to improve data integration services in the cloud (e.g., AWS Glue~\cite{glue}), which provide automated integration capabilities on enterprise data lakes~\cite{vos2022towards}, but currently require end users to manually label examples for entity matching~\cite{glueteaching}. Additionally, zero-shot matchers can be applied for duplicate detection~\cite{hyneslinter} as part of data cleaning in machine learning pipelines~\cite{rein2022,li2021cleanml}, and are also valuable as a primitive for entity alignment in tasks such as table reclamation~\cite{fan2024gen}. 
All of these use cases focus on ingesting and processing large amounts of heterogeneous data from various data sources, such as Excel and CSV files in data lakes, log files or web crawls.
In such settings, schema and type information may often be missing or unreliable.
Therefore, we extend the recently introduced zero-shot setting~\cite{wu2020zeroer} with the {\em additional restriction that no column name or type information is available for the unseen target dataset} (see \Cref{sec:problem} for details), and argue that improving the performance and applicability of zero-shot matchers in this setting is of high practical relevance.

\header{The high cost of LLM-based matchers}  Recently, the promising capabilities of large language models (LLMs)~\cite{fernandez2023large} have led to a resurgence of research on EM~\cite{narayanfm2022,peeters2023entity,zhang2023jellyfish,li2023table}. Approaches such as \texttt{MatchGPT}~\cite{peeters2023entity} or \texttt{TableGPT}~\cite{li2023table} show promising results for zero-shot entity matching by prompting LLMs~\cite{narayanfm2022}. However, these approaches rely on extremely large proprietary models with hundreds of billions or even trillions of parameters~\cite{wang2024mixture}, which require expensive accelerator hardware for deployment. As a result, these approaches incur a hefty cost at a low throughput during inference. The latest commercial LLMs often come with a {\em ``throughput of less than 1~KB per second''}~\cite{liu2024declarative} at a high cost imposed by their pay-per-token model, which results in prices of {\em ``5~USD for processing just 5MB of data''}~\cite{liu2024declarative}. LLM-based entity matchers inherit this high computational cost, which severely limits their scalability and applicability. An indication of this is the fact that even relatively small-scale academic datasets are down-sampled for experimentation with LLM-based approaches~\cite{peeters2023entity}. 

\header{\method{}} We tackle the outlined challenges with our novel matcher \method{}, which is designed for cost-efficient zero-shot entity matching on unseen target data. As illustrated in \Cref{fig:teaser}, \method{} provides competitive prediction quality in zero-shot entity matching at a cost that is more than three orders of magnitude lower than the cost imposed by the latest state-of-the-art matchers, such as \texttt{MatchGPT}~\cite{peeters2023entity} with OpenAI's \texttt{GPT-4} model. In order to achieve this, we model zero-shot entity matching as a sequence classification problem under a transfer learning setup (\Cref{sec:approach}) and fine-tune the small language model \texttt{GPT-2}~\cite{radford2019language} on carefully chosen data for this task (\Cref{sec:approach-datagen}). 
\method{} is easy to apply at inference time and can serve as an out-of-the-box entity matching or deduplication primitive in data science applications, since it neither requires labelled examples nor schema information such as column names or types. 

\vspace{2mm}
\noindent In summary, we provide the following contributions:

\header{Contribution~1 - Model design and data selection (\Cref{sec:approach})} We detail the design of our zero-shot entity matching model \method{}. We model zero-shot entity matching as sequence classification in a transfer learning setup, and {\em fine-tune a small language model} for this task on carefully chosen data samples. As part of our approach, we leverage several novel data selection techniques:

\begin{itemize}[leftmargin=*] 
  \item We use an {\em AutoML filter to identify and include difficult examples}, which often highlight edge cases that the model struggles with. Focusing on them leads to more robust performance.
  \item We {\em augment the fine-tuning data with attribute-level samples} to accommodate for the structural mismatch between text data and relational data without column order.
  \item We propose {\em a heuristic to account for the label imbalance} commonly observed in entity matching data.
\end{itemize}
We conduct an ablation study in \Cref{sec:eval-ablation} to validate all these techniques.
 
\header{Contribution~2 - Competitive prediction quality (\Cref{sec:eval-quality})}  We conduct an extensive evaluation of the prediction quality of our approach against {\em thirteen baselines on nine benchmark datasets}. \method{} achieves the second-highest F1 score overall and outperforms twelve out of thirteen baselines (including LLMs with hundreds of billions of parameters), often by a large margin of more than ten percent in F1 score. This is remarkable, since \method{} has several orders of magnitude less parameters than many LLM-based approaches.

\header{Contribution 3 - High throughput and cost-efficiency (\Cref{sec:eval-performance}) }  We evaluate the computational performance and deployment cost of \method{} in comparison to existing matchers based on large language models. We find that \method{} exhibits attractive performance characteristics: its average prediction quality is within 4.4\% of the state-of-the-art method \texttt{MatchGPT} based on the proprietary trillion-parameter model \texttt{GPT-4}, yet the latter requires four orders of magnitude more parameters and incurs a 3,899 times higher inference cost (in dollars per 1,000 tokens).

\header{Contribution~4 - Model and reproducibility} We make our model, as well as the source code of \method{} and our experiments available at: \textcolor{blue}{\url{https://github.com/Jantory/anymatch}}

\phantom{0000\\0000}

\section{Problem Statement}
\label{sec:problem}
We introduce entity matching and detail the restrictions for our tackled zero-shot setting. Furthermore, we discuss a set of use cases for zero-shot entity matching.

\header{Entity matching} The entity matching problem is to predict whether the pair of records $(r_l, r_r)$ with $r_l \in \mathrm{R}_{\text{left}}$ and $r_r \in \mathrm{R}_{\text{right}}$ refers to the same real-world entity or not. $\mathrm{R}_{\text{left}}$ and $\mathrm{R}_{\text{right}}$ denote two input relations with $k$ aligned attributes $A = \{a_1, \dots, a_k\}$.

Entity matching is often modelled as a binary classification problem with a labelled training set $\mathcal{D}_{\text{train}} \subset \mathrm{R}_{\text{left}} \times \mathrm{R}_{\text{right}} \times \{0,1\}$. State-of-the-art approaches~\cite{li2020deep} featurise the example pairs based on their attribute names $a_1, \dots, a_k$ and the aligned attribute values $v_{l_1} = r_l[a_1]$, $\dots$, $v_{l_k} = r_l[a_k], v_{r_1} = r_r[a_1]$, $\dots$, $v_{r_k} = r_r[a_k]$. Furthermore, the attribute values may be augmented with additional data, e.g., domain information. 
Note that entity matching can also be used for the deduplication of a single relation, when we use pairs of records from this single relation as input. Furthermore, real-world entity matching systems typically first apply a blocking function to the set $R_l \times R_r$ to form smaller candidate sets as input to the matcher. We focus on the matcher itself and present a general approach which can be easily plugged into existing matching systems and applied to blocked candidates sets.

\header{Zero-shot entity matching} In contrast to classical entity matching, we tackle a more challenging setting, referred to as {\em zero-shot entity matching}~\cite{wu2020zeroer}. In particular, we tackle the entity matching problem under the following restrictions:

\vspace{1mm}
\noindent{\underline{Restriction~1 - Unseen target data}: {\em A zero-shot matcher will not have access to labelled example pairs  for the target relations $\mathrm{R}_{\text{left}}$ and $\mathrm{R}_{\text{right}}$, which means there is no training set available for the unseen target data $\mathcal{D}_{\text{target}}$ .} 

\vspace{1mm}
\noindent{\underline{Restriction~2 - Lack of type information}: {\em There is no column name or column type information accessible for the target relations $\mathrm{R}_{\text{left}}$ and $\mathrm{R}_{\text{right}}$. A zero-shot matcher can only enumerate the attribute values $r[a_1]$, $\dots$, $r[a_k]$ of a record~$r$ from the target relations in a string representation.} 

\vspace{1mm} Previous research like \textit{ZeroER}~\cite{wu2020zeroer} already tackles zero-shot entity matching based on Restriction~1 (no training data for the target relations), but still needs column type information to select appropriate similarity functions (violating Restriction~2).

\begin{figure*}[t!]
    \centering
    \begin{subfigure}[T]{0.648\textwidth}    
      \centering
      \includegraphics[width=\columnwidth,page=1]{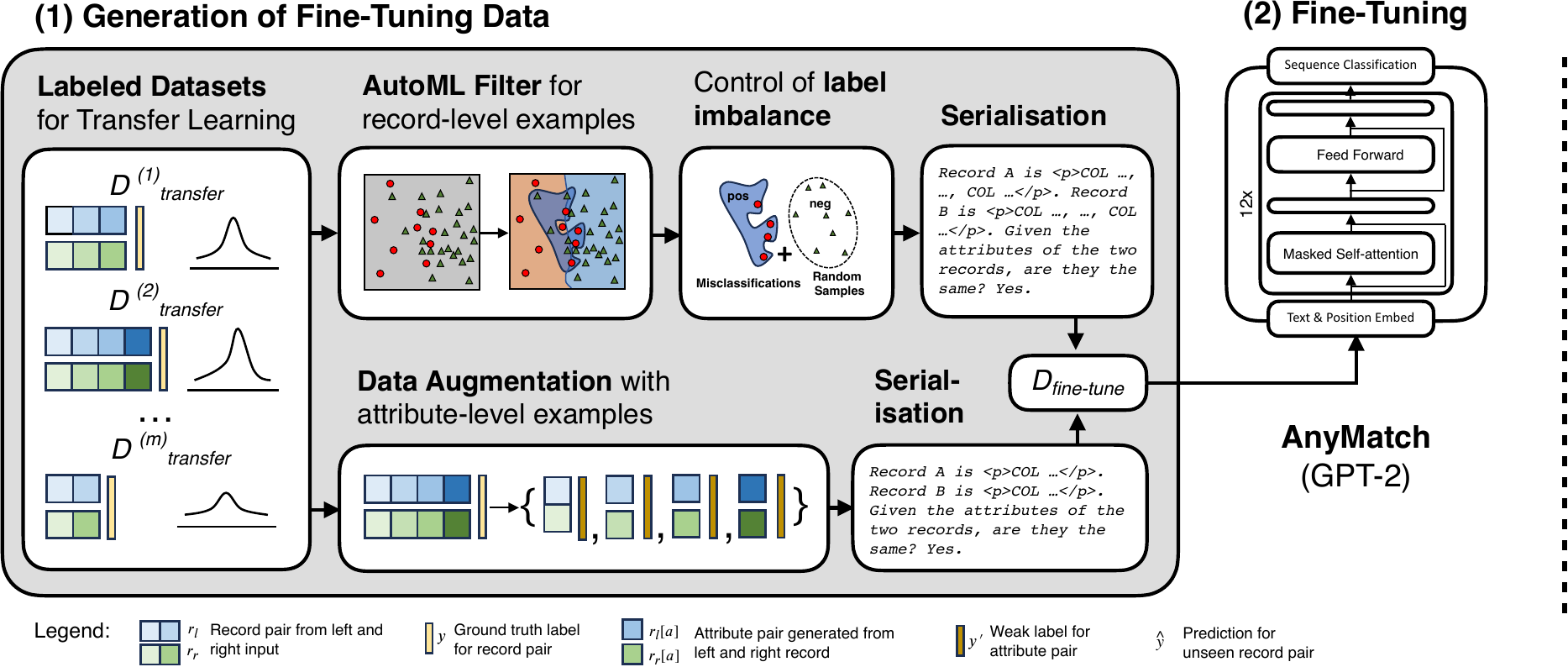}
      \caption{Fine-tuning of \method{} on high quality data generated from the available labeled datasets $\mathcal{D}_{\text{transfer}}^{(1)}, \dots, \mathcal{D}_{\text{transfer}}^{(m)}$ in a transfer learning setup (one-off process).}
      \label{fig:fine-tuning}        
    \end{subfigure}      
    \hfill
    \begin{subfigure}[T]{0.285\textwidth}
      \centering
      \includegraphics[width=\columnwidth,page=2]{anymatch-polished-crop}
      \vspace{0.003cm}
      \caption{Inference on unseen target data $\mathcal{D}_{\text{target}}$ at deployment time.}
      \label{fig:inference}
    \end{subfigure}    
    \caption{High-level overview of \method{} -- (1) We generate fine-tuning data from the available labelled datasets by applying several data selection techniques, and (2)~fine-tune a language model as  zero-shot entity matcher. (3) We use the resulting matcher for inference on unseen target data at deployment time.}
    \label{fig:overview}
\end{figure*}

\header{Use cases} We argue that our proposed zero-shot EM setup is crucial in scenarios where a high level of automation required, and where it is unlikely or impractical to force a domain expert to manually label training data. Furthermore, schema and type information may often be missing or unreliable in these scenarios. Examples for such use cases include {\em data integration services in the cloud}, such as AWS Glue~\cite{glue}, which provide automated integration capabilities on enterprise data lakes. Currently, such services require end users to manually label examples for entity matching~\cite{glueteaching}. They match our restrictions well as they need to be able to ingest and automatically process heterogeneous data from various data sources (e.g., data in relational databases or Excel and CSV files from distributed file systems). Providing competitive matching performance out-of-the-box without labelled examples would greatly improve the applicability of these services. Another use case is {\em duplicate detection and removal in ML pipelines}~\cite{abedjan2016detecting,hyneslinter,rein2022,li2021cleanml}, where the input dataset often originates from various non-relational sources such log files and rarely comes with labelled examples for potential duplicates. Large organisations run hundreds of such pipelines in production~\cite{xin2021production}. Furthermore, zero-shot entity matching is also valuable as {\em a primitive for entity alignment in other data integration tasks}. An example is table reclamation~\cite{fan2024gen}, where a ``fast, approximate instance comparison algorithm'' is required for future work. Note that these use cases depend on cost-efficient matchers and may often require scalable matching on hundreds of thousands or even millions of records.

\section{Approach}
\label{sec:approach}
We give an overview of \method{} in \Cref{sec:approach-transfer}. Next, we detail how we choose our base model and serialization format in \Cref{sec:approach-prompt} and discuss our proposed techniques for selecting the fine-tuning data $\mathcal{D}_{\text{fine-tune}}$ in \Cref{sec:approach-datagen}.

\subsection{Zero-shot Entity Matching as Sequence Classification in a Transfer Learning Setup} 
\label{sec:approach-transfer}

Analogous to existing approaches~\cite{li2020deep}, we treat entity matching as a sequence classification problem, where the input records $r_l$ and $r_r$ are serialised into an ordered sequence $\mathbf{x} = (x_1, \cdots, x_t)$ of $t$~tokens, and a classification model $f: \mathcal{X} \rightarrow \{0,1\}$ predicts whether they refer to the same real-world entity or not. Note that $\mathcal{X}$ denotes the space of possible sequences here. Recall from \Cref{sec:problem} that we aim for a matcher $f$ which can be applied to unseen target $\mathcal{D}_{\text{target}}$ originating from two relations $\mathrm{R}_{\text{left}}$ and $\mathrm{R}_{\text{right}}$, for which no labelled training data $\mathcal{D}_{\text{train}}$ is available.

In order to tackle this challenge, we use a language model as sequence classifier and treat zero-shot entity matching as a {\em transfer learning problem}. We assume that one can access a large set of $m$~labelled datasets $\mathcal{D}_{\text{transfer}}^{(1)}, \cdots, \mathcal{D}_{\text{transfer}}^{(m)}$ from other relations (e.g., all publicly available academic benchmark datasets) to fine-tune a language model as the zero-shot matcher~$f$. Directly concatenating samples from all these available datasets is prone to overfitting, and special care is required to create high quality fine-tuning data. 

\noindent The following steps, illustrated in~\Cref{fig:overview}, detail how to create and use \method{}:
\begin{enumerate}[leftmargin=*]
    \item We generate a single dataset $\mathcal{D}_{\text{fine-tune}}$ from the available labelled datasets $\mathcal{D}_{\text{transfer}}^{(1)}, \cdots, \mathcal{D}_{\text{transfer}}^{(m)}$ by applying several data selection techniques.
    \item We fine-tune a language model on $\mathcal{D}_{\text{fine-tune}}$ as the matcher~$f$.
    \item We use the resulting matcher $f$ in a zero-shot setting for inference on the unseen target data $\mathcal{D}_{\text{target}}$.
\end{enumerate}

\subsection{Base Model and Serialization Format}
\label{sec:approach-prompt}

We leverage the language model \texttt{GPT-2}~\cite{radford2019language}  as our base model. \texttt{GPT-2} is a decoder-only model with 124 million parameters pretrained auto-regressively.
In contrast to encoder-decoder models like \texttt{T5}~\cite{raffel2020exploring} used in~\cite{vos2022towards} with an equivalent number of transformer layers, GPT-2 has no encoder (and therefore only needs about half the number of parameters compared to the 220 million parameters of \texttt{T5-base}). \texttt{GPT-2} is designed for zero-shot tasks and its auto-regressive training enables the model to compress general knowledge and respond to tasks based on the previous context.

\header{Serialisation format} We apply a general, domain-independent serialisation format to turn the pair of tuples to match into a prompt for our base model. Our prompt starts with a representation of both records. The first record $r_l$ with values $v_{l_1}, \dots, v_{l_k}$ is serialised as \texttt{Record A is <p>COL $v_{l_1}$, ..., COL $v_{l_k}$</p>}. The second record $r_r$ with values $v_{r_1}, \dots, v_{r_k}$ is serialised analogously as \texttt{Record B is <p>COL $v_{r_1}$, ..., COL $v_{r_k}$</p>.} and appended. Here, the string \texttt{COL} is used as a marker for a column value, the comma separates column values, while \texttt{<p>} and \texttt{</p>} enclose the record values, to make them more recognisable to the model. Moreover, we represent missing column values with the string \texttt{N/A}. The final question to answer for the model is appended via the following suffix: \texttt{Given the attributes of the two records, are they the same?}.

\header{Example} The record pairs to match can come in any structured form, e.g., as database tuples, from CSV files or in JSON format. We might for example have to make a prediction about records from the music domain representing two different songs from the artist ``David Guetta'':
\begin{Verbatim}[fontsize=\small,commandchars=\\\(\)]
{song_name: (\color(Magenta)"I'm a Machine"), musician: (\color(Magenta)"David Guetta"), 
 price: (\color(Magenta)"$1.29"), release: (\color(Magenta)2011)} 
{title: (\color(MidnightBlue)"Night Of Your Life"), artist: (\color(MidnightBlue)"David Guetta"), 
 price: (\color(MidnightBlue)"$1.29"), year: (\color(MidnightBlue)2011)} 
\end{Verbatim}
\noindent \method{} serialises these two records into the following prompt at prediction time. Note that this prompt format is {\em independent of the data domain} and has {\em no requirement for any schema information} such as column names or column types.
\begin{Verbatim}[fontsize=\small,commandchars=\\\(\)]
Record A is <p>COL (\color(Magenta)I'm a Machine), COL (\color(Magenta)David Guetta), 
COL (\color(Magenta)$1.29), COL (\color(Magenta)2011)</p>. Record B is <p>COL (\color(MidnightBlue)Night Of Your)
(\color(MidnightBlue)Life), COL (\color(MidnightBlue)David Guetta), COL (\color(MidnightBlue)$1.29), COL (\color(MidnightBlue)2011)</p>. Given the
attributes of the two records, are they the same?
\end{Verbatim}

\subsection{Selecting High-Quality Fine-Tuning Data}
\label{sec:approach-datagen}

We now detail the main ideas and motivation for our techniques to turn the set of labelled datasets $\mathcal{D}_{\text{transfer}}^{(1)}, \dots, \mathcal{D}_{\text{transfer}}^{(m)}$ into the data $\mathcal{D}_\text{fine-tune}$ to fine-tune \method{}. We design a set of data selection techniques to make sure that $\mathcal{D}_\text{fine-tune}$ contains high-quality examples that result in a generalisable matcher. Note that the detailed algorithms to generate this data are discussed in the following in \Cref{sec:approach-datagen-algorithms}.

\header{Selection of difficult matching examples via AutoML} A general principle in many EM systems is to categorise candidate pairs based on the difficulty of distinguishing between the represented entities. Unlikely matches are usually pruned early via blocking functions~\cite{papadakis2020blocking}.  Moreover, recent studies~\cite{mudgal2018deep,leone2022critical,papadakis2023critical} indicate that many entity matching benchmark datasets can already be solved using linear models. 

These observations motivate us to think about EM datasets as a mix of record pairs that are easily classified by conventional machine learning models, such as linear models, and another set of pairs that require non-linearity and increased model capacity for accurate prediction. We prioritise the inclusion of the latter, more challenging examples into the fine-tuning data $\mathcal{D}_\text{fine-tune}$ via an {\em AutoML filter}: we train an AutoML model on examples from a labelled dataset $\mathcal{D}_{\text{transfer}}^{(i)}$ and include the positively labelled samples misclassified by the AutoML model in $\mathcal{D}_\text{fine-tune}$. Such samples often highlight the boundaries or edge cases that the model struggles with, and including them can lead to more robust performance and higher transferability.

\header{Augmentation with attribute-level samples} Even though LLMs show promising potential for data wrangling~\cite{narayanfm2022}, there still remain fundamental structural mismatches in applying them to relational data~\cite{badaro2023transformers}, e.g., that relational data has an additional column/attribute structure (which is not present in text) and that the relational model does not impose an order over these attributes. We accommodate for this structural mismatch with an augmentation approach that is inspired by how humans tackle the entity matching task: they will not immediately make decisions at the record level, but will use a combination of comparisons across different attributes. 

In order to account for this observation, we generate several {\em additional attribute-level examples} which only contain pairs of attribute values $(r_l[a], r_r[a])$ for each attribute $a$ from a labeled record pair $(r_l, r_r)$ in $\mathcal{D}_{\text{transfer}}^{(i)}$, and augment the fine-tuning data $\mathcal{D}_\text{fine-tune}$ with these additional samples. 

\header{Controlling label imbalance} In general, the candidate set of record pairs for a matcher is generated through the Cartesian product of the two involved relations (filtered by blocking functions). This typically results in significantly more negatively labelled (non-matching) record pairs than positively labelled matching pairs. Directly using all such entity pairs to fine-tune a model could lead to serious overfitting (e.g., via instance memorisation) on the minority class of positively labelled samples. Moreover, since we make no assumptions about the label distribution in the target data $\mathcal{D}_{\text{target}}$, it is crucial for the final model to perform well on both positive and negative samples.

We address this label imbalance with a heuristic to control the ratio of negative to positive samples and include twice as many negative as positive examples from each dataset $\mathcal{D}_{\text{transfer}}^{(i)}$ in the fine-tuning data $\mathcal{D}_\text{fine-tune}$. The intuition behind this rate is as follows: given a positive pair $(r_l, r_r)$, the model should be able to make different predictions for $(\tilde r_l, r_r)$ and $(r_l, \tilde r_r)$, where $\tilde r_l$ and $\tilde V_r$ are similar but distinct entities from the two tables. Given that such negative samples may not have been annotated with golden labels in real datasets, which have golden label from the data annotator, this ratio serves as a practical measure to ensure a consistent balance between positive and negative pairs.

\subsection{Data Selection Algorithms}
\label{sec:approach-datagen-algorithms}

We now detail the concrete algorithms to turn the set of labelled transfer datasets $\mathcal{D}_{\text{transfer}}^{(1)}, \dots, \mathcal{D}_{\text{transfer}}^{(m)}$ into the data $\mathcal{D}_\text{fine-tune}$ which we use for fine-tuning. 
The function \textsc{finetune\_anymatch} in \Cref{alg:training-anymatch} shows the high-level procedure as outlined previously: we first generate record-level examples (\Cref{alg:training-data-generation-recordlevel}) which are then augmented with attribute-level examples (\Cref{alg:training-data-generation-attributelevel}), which we finally use to fine-tune our base model \texttt{GPT-2}.

{\small
\begin{algorithm}[ht!]
\caption{Fine-tuning \method{} via transfer learning.}\label{alg:training-anymatch}
\begin{algorithmic}[1]
\Function{finetune\_anymatch}{$\mathcal{D}^{(1)}_{\text{transfer}}, \dots, \mathcal{D}^{(m)}_{\text{transfer}}, n_a, n_r$}    \State $\mathcal{D}_{\text{fine-tune}} \leftarrow \emptyset$
    \Statex \textcolor{gray}{\phantom{00}-- Generate record-level samples}
    \For{$i \in 1 \dots m$}
        \State $S_\text{record} \leftarrow \textsc{gen\_recordlevel}(\mathcal{D}^{(i)}_{\text{transfer}}, n_r)$
        \State $\mathcal{D}_{\text{fine-tune}} \leftarrow \mathcal{D}_{\text{fine-tune}} \cup S_\text{record}$        
    \EndFor
    \Statex \textcolor{gray}{\phantom{00}-- Generate attribute-level samples}    
    \State $S_\text{attribute} \leftarrow \textsc{gen\_attributelevel}(\mathcal{D}^{(1)}_{\text{transfer}}, \dots, \mathcal{D}^{(m)}_{\text{transfer}}, n_a)$    
        \State $\mathcal{D}_{\text{fine-tune}} \leftarrow \mathcal{D}_{\text{fine-tune}} \cup S_\text{attribute} $            
    \Statex \textcolor{gray}{\phantom{00}-- Fine-tune a small language model for EM}    
    \State $f \leftarrow$ fine-tune GPT-2 on $\mathcal{D}_{\text{fine-tune}}$
    \State \Return $f$
\EndFunction
\end{algorithmic}
\end{algorithm}}

{\small
\begin{algorithm}[ht!]
\caption{Generating record-level fine-tuning samples.}\label{alg:training-data-generation-recordlevel}
\begin{algorithmic}[1]
\Function{gen\_recordlevel}{$\mathcal{D}_{\text{transfer}}, n_r$}    
    \State $S_\text{record} \leftarrow \emptyset$ 
        \If{$|\mathcal{D}_{\text{transfer}}| > n_r$}
            \Statex \textcolor{gray}{\phantom{0000}-- AutoML filter to select difficult examples}        
            \State $f_\text{auto} \leftarrow$ train AutoML classifier on $\mathcal{D}_{\text{transfer}}$ \label{alg:line-automl-start}
            \State $D^+_\text{wrong} \leftarrow$ matching pairs misclassified by $f_\text{auto}$ \label{alg:line-automl-end}  
            \Statex \textcolor{gray}{\phantom{0000}-- Downsampling to control label imbalance}    
            \State $n^+_\text{wrong} \leftarrow |D^+_\text{wrong}|$ \label{alg:line-sampling-matches-start}            
            \State $n^+_{D} \leftarrow$ number of matching pairs in $\mathcal{D}_{\text{transfer}}$ 
            \State $n_p \leftarrow \frac{1}{3} n_r$            
            \State $n^+ \leftarrow \min(n^+_{D}, n_p)$
            \If{$n^+_\text{wrong} \geq n^+$}
                \State $D^+ \leftarrow$ sample $n^+$ pairs from $D^+_\text{wrong}$
            \Else
                \State $D^+_\text{corr} \leftarrow$ matching pairs correctly classified by $f_\text{auto}$             
                \State $D^+_\text{corr} \leftarrow$ sample $(n^+ - n^+_\text{wrong})$ pairs from $D^+_\text{corr}$            
                \State $D^+ \leftarrow D^+_\text{wrong} \cup D^+_\text{corr}$ \label{alg:line-sampling-matches-end} 
            \EndIf
            \State $D^- \leftarrow$ sample $2n^+$ non-matching pairs from $\mathcal{D}_{\text{transfer}}$         
            \State $D^\pm \leftarrow D^+ \cup D^-$
        \Else
            \Statex \textcolor{gray}{\phantom{0000}-- No filtering for tiny datasets}      
            \State $D^\pm \leftarrow \mathcal{D}_{\text{transfer}}$ \label{alg:line-nosampling}                         
        \EndIf
        \Statex \textcolor{gray}{\phantom{00}-- Generation of serialised samples}        
         \For{$(r_l, r_r, y) \in D^\pm$} \label{alg:line-record-level-instance-start}
             \State $V_l \leftarrow \textsc{enumerate\_attribute\_values}(r_l)$             
             \State $V_r \leftarrow \textsc{enumerate\_attribute\_values}(r_r)$
             \State $S_\text{record} \leftarrow S_\text{record} \cup \textsc{serialize}(V_l, V_r, y)$
             \Statex \textcolor{gray}{\phantom{0000}-- Include ``flipped'' sample}             
             \State $S_\text{record} \leftarrow S_\text{record} \cup \textsc{serialize}(V_r, V_l, y)$ \label{alg:line-record-level-instance-end}             
         \EndFor          
    \State \Return $S_\text{record}$
\EndFunction
\end{algorithmic}
\end{algorithm}}

\header{Record-level sample generation with AutoML selection} We generate record-level samples independently from each dataset $\mathcal{D}_{\text{transfer}} \in \mathcal{D}^{(1)}_{\text{transfer}}, \dots, \mathcal{D}^{(m)}_{\text{transfer}} $ as described in \Cref{alg:training-data-generation-recordlevel}. On a high-level, we aim to resample each dataset $\mathcal{D}_{\text{transfer}}$ to have a ratio of twice as much non-matching as matching record pairs. Furthermore, we focus on identifying and including difficult matching pairs in the fine-tuning data generated from $\mathcal{D}_{\text{transfer}}$. 

We generate $n_r$ samples from each dataset $\mathcal{D}_{\text{transfer}}$ as follows. First, note that $n_r$ is a hyperparameter and the we use the full dataset $\mathcal{D}_{\text{transfer}}$ if it has less pairs (Line~\ref{alg:line-nosampling}). We detect difficult matching pairs by training an AutoML classifier $f_\text{auto}$ on $\mathcal{D}_{\text{transfer}}$ and identify the set $D^+_\text{wrong}$ of matching record pairs that were misclassified by $f_\text{auto}$ as false negatives (Lines~\ref{alg:line-automl-start}-\ref{alg:line-automl-end}). We leverage these misclassified samples to create the set of matching pairs to use for fine-tuning $D^+$, by sampling $\frac{1}{3}n_r$ difficult matching pairs from $D^+_\text{wrong}$  (and additionally include correctly classified pairs as well if there are not enough difficult pairs, Lines~\ref{alg:line-sampling-matches-start}-\ref{alg:line-sampling-matches-end}). Next, we randomly sample twice as many non-matching pairs from $\mathcal{D}_{\text{transfer}}$ to form the set of non-matching pairs $D^-$. Finally, we combine $D^+$ and $D^-$ to form the set of pairs $D^\pm$ and turn each retained labelled record pair $(r_l, r_r, y)$ from $D^\pm$ into two instances based on our serialisation format (Lines~\ref{alg:line-record-level-instance-start}-\ref{alg:line-record-level-instance-end}). We generate a sample based on the value pairs $(V_l, V_r)$ and a second sample where the order of the value pairs is flipped: $(V_r, V_l)$. At this level, all attribute values $V_l$ from the left record $r_l$ and $V_r$ from the right record $r_r$ are included in the serialisation.

{\small
\begin{algorithm}[ht!]
\caption{Generating attribute-level fine-tuning samples.}\label{alg:training-data-generation-attributelevel}
\begin{algorithmic}[1]
\Function{gen\_attributelevel}{$\mathcal{D}^{(1)}_{\text{transfer}}, \dots, \mathcal{D}^{(m)}_{\text{transfer}}, n_a$}    
    \State $S_\text{attribute} \leftarrow \emptyset$     
    \Statex \textcolor{gray}{\phantom{00}-- Identify all possible attributes}          
    \State $A \leftarrow$ all attributes appearing in $\mathcal{D}^{(1)}_{\text{transfer}}, \dots, \mathcal{D}^{(m)}_{\text{transfer}}$
    \Statex \textcolor{gray}{\phantom{00}-- Collect all attribute pairs}      
    \For{attribute $a_i \in A$}      
        \State $G \leftarrow \emptyset$    \label{alg:line-attribute-grouping-start}
        \For{$j \in 1 \dots m$}      
            \If{$a_i$ in attributes of $\mathcal{D}^{(j)}_{\text{transfer}}$}
                \For{$(r_l, r_r, y) \in \mathcal{D}^{(j)}_{\text{transfer}}$}
                    \State $v_l \leftarrow r_l[a_i]$
                    \State $v_r \leftarrow r_r[a_i]$
                    \State $G \leftarrow G \cup (v_l, v_r, y)$ \label{alg:line-attribute-grouping-end}
                \EndFor        
            \EndIf
        \EndFor
        \Statex \textcolor{gray}{\phantom{0000}-- Balance the pairs per attribute}          
        \State $n^+ \leftarrow$ number of matching pairs in $G$ \label{alg:line-attribute-balancing-start}
        \State $n^- \leftarrow$ number of non-matching pairs in $G$
        \State $n_{\text{balance}} \leftarrow \min(|n^+|, |n^-|)$        
        \State $G^+ \leftarrow$ $n_{\text{balance}}$ matching pairs sampled from $G$
        \State $G^- \leftarrow$ $n_{\text{balance}}$ non-matching pairs sampled from $G$    
        \State $G^\pm \leftarrow P^+ \cup P^-$        
        \Statex \textcolor{gray}{\phantom{0000}-- Down-sample the pairs per attribute}                  
        \If{$|G^\pm| > n_{\text{a}}$}
            \State $G^\pm \leftarrow$ randomly sample $n_{\text{a}}$ pairs from $P^\pm$ \label{alg:line-attribute-balancing-end}      
        \EndIf     
        \Statex \textcolor{gray}{\phantom{0000}-- Generate attribute-level samples}                  
        \For{$(v_l, v_r, y) \in P^\pm$}
            \State $S_\text{attribute} \leftarrow S_\text{attribute} \cup \textsc{serialize}(v_l, v_r, y)$ \label{alg:line-attribute-gen}
        \EndFor        
    \EndFor 
    \State \Return $S_\text{attribute}$
\EndFunction
\end{algorithmic}
\end{algorithm}}

\header{Augmentation with attribute-level samples} In addition to the record-level samples, we also generate attribute-level samples for fine-tuning, as previously discussed in \Cref{sec:approach-datagen}. We detail their generation process in \Cref{alg:training-data-generation-attributelevel}. The goal of this function is to generate samples\footnote{Note that this requires us to have access to attribute names at data generation time, however the actual attribute names are not part of the generated fine-tuning data and thereby not observed by our model.} for all attributes $A$ occurring in any of the datasets $\mathcal{D}^{(0)}_{\text{transfer}}, \dots, \mathcal{D}^{(m)}_{\text{transfer}}$. Therefore, the data access pattern is different than before. Instead of processing each dataset $D^{(j)}_{\text{transfer}}$ independently, we now group the data by attribute (Lines~\ref{alg:line-attribute-grouping-start}-\ref{alg:line-attribute-grouping-end}) and process each record group $G$ containing all labelled pairs of values of a particular attribute $A_i$ independently. We balance each group $G$ to contain positively and negatively labelled attribute pairs in equal proportions and make sure that we use at most $n_a$ pairs per attribute (Lines~\ref{alg:line-attribute-balancing-start}-\ref{alg:line-attribute-balancing-end}). Note that the threshold $n_a$ is a hyperparameter again. Recall that we use a ratio of one to two for record-level samples, but we do not apply the same ratio for attribute-level samples. This is because attribute-level samples are generated using simple heuristic rules without ground-truth labels on attribute-level matches. Therefore, we use a balanced ratio between positive and negative samples to minimise inductive bias. Finally, we turn each retained labelled attribute value pair into a sample based on our serialisation format, using only a single attribute value $v_l$ from the left record $r_l$ and the corresponding single attribute value $v_r$ from the right record~$r_r$ (\Cref{alg:line-attribute-gen}).

\header{Example} We detail the data generation using a toy example. The transfer learning data for \method{} might contain the following matching record pair from the music domain for the same song from the artist ``Drake'':
\begin{Verbatim}[fontsize=\small,commandchars=\\\(\)]
{song_name: (\color(Magenta)"6PM In New York"), artist: (\color(Magenta)"Drake"), 
 genre: (\color(Magenta)"Hip-Hop/Rap Music - Hardcore Rap - Rap - R&B/Soul -) 
 (\color(Magenta)Contemporary R&B"), price: (\color(Magenta)"$1.29"), released: (\color(Magenta)"13-Feb-15")}
{title: (\color(MidnightBlue)"6PM In New York [Explicit]"), musician: (\color(MidnightBlue)"Drake"), 
 genre: (\color(MidnightBlue)"Rap & Hip-Hop"), price: (\color(MidnightBlue)"$ 1.29"), 
 release_date: (\color(MidnightBlue)"February 13 2015")}
\end{Verbatim}
This may be considered a difficult pair since only the artist name is an exact match, so it might be selected by our AutoML filter, which identifies difficult positive examples. In that case, we will first generate a record-level instance from the record pair with all attribute values, based on our serialization format:
\begin{Verbatim}[fontsize=\small,commandchars=\\\(\)]
Record A is <p>COL (\color(Magenta)6PM In New York), COL (\color(Magenta)Drake), COL (\color(Magenta)Hip-Hop/)
(\color(Magenta)Rap Music - Hardcore Rap - Rap R&B/Soul - Contemporary R&B), 
COL (\color(Magenta)$1.29), COL (\color(Magenta)13-Feb-15)</p>. Record B is <p>COL (\color(MidnightBlue)6PM In New) 
(\color(MidnightBlue)York [Explicit]), COL (\color(MidnightBlue)Drake), COL (\color(MidnightBlue)Rap & Hip-Hop), COL (\color(MidnightBlue)$ 1.29), 
COL (\color(MidnightBlue)February 13 2015)</p>. Given the attributes of the two 
records, are they the same? Yes.
\end{Verbatim}

Our approach will also generate another instance where the order of the records is flipped, and  additional examples which only use a single pair of attribute values for augmentation with attribute-level data:
\begin{Verbatim}[fontsize=\small,commandchars=\\\(\)]
Record A is <p>COL (\color(Magenta)6PM In New York)</p>. Record B is
<p>COL (\color(MidnightBlue)6PM In New York [Explicit])</p>. Given the attributes 
of the two records, are they the same? Yes.
\end{Verbatim}

\begin{Verbatim}[fontsize=\small,commandchars=\\\(\)]
Record A is <p>COL (\color(Magenta)Hip-Hop/Rap Music - Hardcore Rap - Rap) 
(\color(Magenta)R&B/Soul - Contemporary R&B)</p>. Record B is <p>COL (\color(MidnightBlue)Rap &)
(\color(MidnightBlue)Hip-Hop)</p>. Given the attributes of the two records, are 
they the same? Yes.
\end{Verbatim}

\begin{Verbatim}[fontsize=\small,commandchars=\\\(\)]
Record A is <p>COL (\color(Magenta)February 13 2015)</p>. Record B is
<p>COL (\color(MidnightBlue)13-Feb-15)</p>. Given the attributes of the two 
records, are they the same? Yes.
\end{Verbatim}

\section{Implementation}
\label{sec:implementation}

We implement \method{} based on the existing \texttt{GPT2Tokenizer} and ~\texttt{GPT2ForSequenceClassification} classes in PyTorch~\cite{paszke2019pytorch} for our chosen base model \texttt{GPT-2}~\cite{radford2019language} from the \texttt{transformers} library. 

We implement the AutoML filter, which selects difficult training examples, based on the \texttt{TabularPredictor} from Amazon's {\em AutoGluon} library~\cite{erickson2020autogluon}. When filtering individual datasets, we convert them into a \texttt{TabularDataset} instance and subsequently construct a \texttt{TabularPredictor} by specifying the name of the label column that indicates matches. We do not constrain the time budget for the AutoML training (e.g., we do not set a \texttt{time\_limit} argument in the \texttt{fit()} method), but observe that the fitting procedure can be completed within minutes for the datasets we are using.

We fine-tune \texttt{GPT-2} via stochastic gradient descent using Adam as optimiser with a fixed learning rate of $0.00002$ and weight decay of 0.01. We run fine-tuning for 50 epochs during which we keep track of the best model by evaluating the F1 score on held-out validation data. We apply early stopping if we observe no significant performance improvement on the validation set for six epochs. 

\section{Related Work}
\label{sec:related}
We summarise existing entity matching approaches (both with and without zero-shot capabilities).

\header{Entity matching approaches applicable in the zero-shot setting} We start by discussing approaches that are applicable in the zero-shot setting which is in the focus of our paper. 

A seminal work in this area is {\em ZeroER}~\cite{wu2020zeroer}, which is a parameter-less method, explicitly designed for the zero-shot case without any training data for the target dataset. The approach is built on the observation that the similarity vectors for matching records are distributed differently than the similarity vectors for non-matching records. In contrast to our approach, there are several drawbacks though: the method requires information about the types of the columns and a decision on the similarity functions to use, is only applicable in a batch setting and cannot match single record pairs in isolation (which for example makes debugging false predictions difficult), and relies on distributional assumptions which may not hold on every dataset.

{\em Ditto}~\cite{li2020deep} is a state-of-the-art entity matching approach, based on fine-tuning a Bert encoder~\cite{devlin2018bert}. Ditto injects domain knowledge into the data during serialisation and augments the training data to enhance the model's ability to differentiate between challenging entity pairs. This process includes actions such as dropping columns and editing spans of tokens. Since Ditto models do not rely on a hard-coded schema during training, they can also be applied to unseen target datasets with a different schema in a zero-shot setting. In a similar direction, \cite{zhangdirections} recently proposed a vision to leverage a selection of LoRA-tuned domain-specific models for entity matching.

{\em Jellyfish}~\cite{zhang2023jellyfish} is a general LLM-based approach targeting four data preprocessing tasks (including entity matching). Jellyfish leverages two LLaMA2-13B models, which are fine-tuned in an instruction tuning fashion. The corresponding data is specifically created to accommodate multiple data preprocessing tasks. Essentially, one LLaMA model is tasked with classification, providing detailed reasoning, while the second model interprets this output to refine the reasoning process further. Jellyfish is explicitly designed to tackle zero-shot data preparation scenarios on unseen datasets.

Narayan et al. \cite{narayanfm2022} showed that prompting large commercial LLMs with serialised records (and carefully chosen few-shot examples) can lead to competitive matching performance. {\em MatchGPT}~\cite{peeters2023entity} enhances the chosen prompts and evaluates a wide variety of base models and prompt formats for both zero-shot and few-shot entity matching. {\em TableGPT}~\cite{li2023table} applies a similar approach, but enhances the LLMs with ``table fine-tuning'' to teach them various data preparation tasks. This approach is designed for both zero-shot and few-shot scenarios.

\header{Approaches without zero-shot capabilities} Next, we discuss a set of methods for entity matching, which are not able to adapt to the zero-shot setting on unseen data, since their feature encoding depends on the schema of the target data to match. Magellan~\cite{doan2020magellan} focuses on building an end-to-end system for entity matching, based on classical machine learning methods. {\em GNEM}~\cite{chen2021gnem} employs a graph neural network approach to entity matching, where each node represents an entity pair and encodes semantic information and interactions. {\em HierMatch}~\cite{fu2021hierarchical} introduces a novel approach to entity matching by constructing a hierarchical structure that progresses from the token level to the attribute level, and finally to the entity level. {\em DeepMatcher}~\cite{xie2024deepmatcher} is a transformer-based neural network for entity matching with various components to improve classification performance across three distinct types of input: structured, text, and dirty data. {\em MCAN}~\cite{zhang2020multi} extends the DeepMatcher model with an attention mechanism after the attribute matching phase. 

\header{Orthogonal aspects} In addition to research on improving the performance of matchers, there is also important work on orthogonal aspects of the entity matching problem. A recent study~\cite{shahbazi2023through} on the fairness of entity matchers uncovered problems in light of the coverage of certain demographic groups or with the similarity characteristics of certain names. The REIN~\cite{rein2022} and CleanML~\cite{li2021cleanml} benchmarks investigate the impact of data errors and cleaning techniques (including entity matching in the form of deduplication) on the downstream performance of ML models. Furthermore, there are several critical assessments of the difficulty of the entity matching task~\cite{papadakis2023critical,leone2022critical,mudgal2018deep}. We interpret their findings as evidence for the potential of zero-shot matchers.

\section{Experimental Evaluation}
In the following we evaluate the prediction quality of \method{} in comparison to several baselines on a variety of entity matching datasets in \Cref{sec:eval-quality}. We additionally investigate the computational performance and deployment cost of our method in \Cref{sec:eval-performance}. Finally, we conduct an ablation study to validate our model design and our choices for the fine-tuning data generation in \Cref{sec:eval-ablation}. The source code of our experiments is available at \textcolor{blue}{\url{https://github.com/Jantory/anymatch}}.

\subsection{Prediction Quality}
\label{sec:eval-quality}

In our first experiment, we measure the prediction quality of \method{} for zero-shot entity matching on  nine benchmark datasets. Our goal is to confirm that our method offers an attractive trade-off between model size and prediction quality.

\header{Datasets} We experiment on nine benchmark datasets detailed in \Cref{tab:datasets}, which are commonly used in entity matching research \cite{li2020deep,narayanfm2022,peeters2023entity}.
\begin{table}[b!]
  \centering
\begin{tabular}{l l l | r }
\toprule
\textbf{Acronym} & \textbf{Dataset} & \textbf{Domain} & \textbf{\#Samples}\\
\midrule
\texttt{ABT} & Abt-Buy & web products & 8,865\\
\texttt{AMGO} & Amazon-Google & software &  11,460\\
\texttt{BEER} & Beer & food & 450\\
\texttt{DBAC} & DBLP-ACM & citation & 12,363\\
\texttt{DBGO} & DBLP-Google & citation & 28,707\\
\texttt{FOZA} & Fodors-Zagats & food & 946\\
\texttt{ITAM} & iTunes-Amazon & music & 539\\
\texttt{WAAM} & Walmart-Amazon & electronics & 10,242\\
\texttt{WDC} & WDC & web products & 6,239\\
\bottomrule
\end{tabular}
  \caption{Benchmark datasets from \cite{narayanfm2022} and \cite{peeters2023entity} with their corresponding domain.}
  \label{tab:datasets}  
\end{table}
Our most important baseline (with the highest performance for zero-shot EM) is \texttt{MatchGPT}~\cite{peeters2023entity}, which leverages commercial LLM APIs. Due to the high costs of these APIs, the \texttt{MatchGPT} study down-samples a test set if it exceeds 1,250 samples (and reduces it to at most 250 positive samples and 1,000 negative samples). We adapt their metholodogy and consult \texttt{MatchGPT}'s code and data repository\footnote{\url{https://github.com/wbsg-uni-mannheim/MatchGPT/tree/main/LLMForEM}} to make sure that our test sets contain exactly the same record pairs.

\header{Metrics} In line with existing research, we report the F1 score for the predictions on the entity matching tasks. 

\header{Setup for \method{}} Evaluating models for our zero-shot setting is challenging, since example data for transfer learning is required.
We evaluate \method{} (and related models like \texttt{Ditto}) with a methodology we refer to as ``leave one dataset out'' fine-tuning: to test on a given unseen target dataset, we use training data from all the other eight datasets as transfer data $\mathcal{D}_{\text{transfer}}^{(1)}, \dots, \mathcal{D}_{\text{transfer}}^{(8)}$ to generate our fine-tuning data (as detailed in \Cref{sec:approach-datagen}).
E.g., to evaluate on \texttt{FOZA}, we use the remaining eight datasets' training data for transfer learning (\texttt{ABT}, \texttt{AMGO}, \texttt{BEER}, \texttt{DBAC}, \texttt{DBGO}, \texttt{ITAM}, \texttt{WAAM} and \texttt{WDC}). This methodology conforms to our zero-shot restrictions (\Cref{sec:problem}) since no examples from the target dataset are fine-tuned on. We make all variants of our trained models available as part of our supplemental material.

\headerl{Hyperparameters} We use a fixed set of hyperparameters across all evaluations. We set the number of record-level instances to generate per dataset $n_r = 1,200$ and the number of attribute-level instances to generate per attribute $n_a = 600$. When we fine-tune \method{}, we use a learning rate of $0.00002$ and always choose the largest batch size that fits into the memory of the underlying GPU.

\header{Baselines} We compare \method{} against the following thirteen baselines and exclude methods which are not applicable in the zero-shot setting (and refer to \Cref{sec:related} for a detailed discussion of the corresponding approaches).

\headeru{Trainingless baselines} We include the following baselines, which do not require any labelled example data.

\begin{itemize}[leftmargin=*]
  \item{} \texttt{StringSim} -- a trivial baseline method, which serialises both tuples to compare by casting each column to a string and concatenating the values with a comma separator. Next, this method computes the string similarity of the serialised tuples via the Ratcliff/Obershelp algorithm from Python's \texttt{difflib} package and predicts a match if the corresponding similarity is greater than 0.5.
  \item{} \texttt{ZeroER} \cite{wu2020zeroer} -- A seminal parameterless zero-shot approach to entity matching, based on distributional assumptions about the similarities of matching and non-matching pairs. We leverage the implementation\footnote{\url{https://github.com/mohamedyd/rein-benchmark/tree/master/cleaners/zeroer}} provided in the REIN benchmark~\cite{rein2022}. Note that \texttt{ZeroER} requires knowledge about the column types to choose the corresponding similarity metrics to compute, which partially violates our zero-shot setting from \Cref{sec:problem}.
\end{itemize}

\begin{table*}[t!] 
\centering
\resizebox{\linewidth}{!}{%
\begin{tabular}{l @{\hskip 0.3in} r @{\hskip 0.3in} rrrrrrrrr @{\hskip 0.3in} r}
\toprule
 & \#params & \multicolumn{9}{c}{\textbf{Unseen Target Dataset}} &   \\
\textbf{} & (millions) & \texttt{ABT} & \texttt{AMGO} & \texttt{BEER} & \texttt{DBAC} & \texttt{DBGO} & \texttt{FOZA} & \texttt{ITAM} & \texttt{WAAM} & \texttt{WDC} & Mean \\
\midrule

\texttt{StringSim} & - & 32.19& 36.43& 29.55& 73.94& 60.10& 22.00 & 51.28& 27.98& 28.46& 40.21\\
\texttt{ZeroER} & - & 54.13 & 45.79 & 67.67 & 93.31 & 85.15 & \textbf{100.00} & 70.12 & 43.12 & 42.46 & 66.86\\

\texttt{Ditto} & 110 &53.94& 50.28& 73.68& \underline{95.22}& 87.79& 69.84& 72.97& 39.87& 50.86& 66.05\\

\texttt{Jellyfish} & 13,000 & 81.13 & (59.72) & (80.00) & (98.40) & (92.65) & (97.67) & (82.61)& 65.78 & 42.55 & (77.83) \\

\texttt{MatchGPT [Mixtral-8x7B]} & 56,000 & 79.02& 31.65& 70.91& 87.63& 66.15& 88.23& 62.13& 50.56& 42.04& 64.26 \\
\texttt{MatchGPT [SOLAR]} & 70,000 & 85.04& 19.01& 71.23& 88.60& 52.05& 89.37& 61.20& 65.32& 56.52& 65.37 \\

\texttt{MatchGPT [Beluga2]} & 70,000 & 83.51& 42.01& 76.34& 92.43& 72.68& 82.17& 54.36& 63.46& 54.97& 69.10 \\

\texttt{GPT-3} & 175,000 & n/a& 54.3\phantom{0} & 78.6\phantom{0}& 93.5\phantom{0} & 64.6\phantom{0} & 87.2\phantom{0} & 65.9\phantom{0} & 60.6\phantom{0} & n/a & n/a \\

\texttt{TableGPT [GPT-3.5-text-davinci-002]} & 175,000 & n/a & \underline{65.7}\phantom{0} & 72.7\phantom{0} & 84.7\phantom{0} & 86.1\phantom{0} & 87.2\phantom{0} & 78.8\phantom{0} & \textbf{69.1}\phantom{0} & n/a& n/a\\
\texttt{TableGPT [GPT-3.5-text-chat-davinci-002]} & 175,000 & n/a & 56.6\phantom{0} & \underline{92.3}\phantom{0} & 93.2\phantom{0} & \textbf{91.1}\phantom{0} & \textbf{100.00} & \underline{86.2}\phantom{0} & 67.8\phantom{0} & n/a & n/a \\

\texttt{MatchGPT [GPT-3.5-Turbo03]} & 175,000 & 74.32& 57.91& 78.78& 88.73& 79.15& 82.35& 56.10& 64.26& \underline{76.51}& 73.12 \\
\texttt{MatchGPT [GPT-3.5-Turbo06]} & 175,000 & 82.30& 44.06& 70.00& 93.28& 78.22& 93.02& 57.57& 68.77& 60.62& 71.98 \\
\texttt{MatchGPT [GPT-4]} & 1,760,000 & \textbf{94.40}& \textbf{74.91}& 69.57& \textbf{95.60}& 87.22& \underline{97.67}& 82.35 & \textbf{89.67}& \textbf{85.83}& \textbf{86.36}  \\

\midrule 

\method{} (ours) & 124 & \underline{86.05}& 55.08& \textbf{96.55}& 93.61& \underline{90.59}& \textbf{100.00}& \textbf{90.91}& 61.51& 63.31& \underline{81.96}\\
\bottomrule
\end{tabular}}
\caption{F1 scores for zero-shot entity matching. (Best score in bold, second-best score underlined, numbers in brackets indicate that the model has seen data from a dataset at training time). Our proposed method \method{} achieves the second-highest score of 81.96 and outperforms all baselines except \texttt{MatchGPT} with the proprietary trillion-parameter model \texttt{GPT-4}, which has four orders of magnitude more parameters.}
\label{tab:zm_label}
\end{table*}

\headeru{Small language models for entity matching} We include two additional baselines leveraging smaller language models. 
\begin{itemize}[leftmargin=*]
  \item{} \texttt{Ditto} \cite{li2020deep} -- A deep entity matching model based on the \texttt{Bert} language model~\cite{devlin2018bert} with 110 million parameters. We evaluate this model in the zero-shot setting with ``leave one dataset out''-fine-tuning, analogous to \method{}. We leverage the original code from the authors' Github repository, but configure the model to not use the optimisation which adds ``domain knowledge [...] highlighting important pieces of the input'', since such knowledge is not available in a zero-shot setting.
  \item \texttt{Jellyfish} \cite{zhang2023jellyfish} -- We leverage the publicly available pretrained 13~billion parameter model and prompt format provided by the authors, which naturally supports zero-shot entity matching. Unfortunately, the authors used six out of the nine benchmark datasets during the multi-task training of their model. As a consequence, we cannot fairly evaluate the model in a zero-shot setting on this data, since it has already seen training data for these tasks. We still report the resulting numbers for completeness, but put them in brackets to indicate that they do not originate from a zero-shot setup.
 \end{itemize} 

\headeru{Entity matching with large language models} We include three baselines which leverage open and commercial large language models with up to a trillion parameters for the entity matching task.
 \begin{itemize}[leftmargin=*] 
  \item \texttt{GPT-3} -- A popular commercial LLM with 175 billion parameters from OpenAI~\cite{brown2020language}. We report the numbers from \cite{narayanfm2022} for their zero-shot prompt setting. Unfortunately, these numbers are only available for the full versions of seven out of our nine datasets, and we cannot compute the numbers for other two datasets since \texttt{GPT-3} not available anymore. We nevertheless include this baseline for completeness.
  \item \texttt{MatchGPT} -- This approach from \cite{peeters2023entity} prompts large language models for the entity matching task. We leverage zero-shot prompting based on the \texttt{general-complex-force} prompt format from \cite{peeters2023entity}, which showed the best performance without domain-specific information in the cited study. We evaluate the three variants \texttt{MatchGPT~[Mixtral]}, \texttt{MatchGPT~[SOLAR]} and \texttt{MatchGPT [Beluga2]} which are based a set of large open-weight models ranging from 56 billion to 70 billion parameters. Furthermore, we include the variants \texttt{MatchGPT~[GPT-3.5-Turbo03]}, \texttt{MatchGPT~[GPT-3.5-Turbo06]} and \texttt{MatchGPT~[GPT-4]} which are based on three commercial LLMs from OpenAI. The parameter sizes of these models are assumed to be 175 billion for \texttt{GPT-3.5} and 1.76 trillion (8x220B)~\cite{wang2024mixture} for \texttt{GPT-4}.
  We report the F1 scores from \cite{peeters2023entity} for six out of the nine datasets covered in their study and compute the scores for the three remaining datasets not covered in \cite{peeters2023entity} ourselves (\texttt{BEER}, \texttt{FOZA}, \texttt{ITAM}).

  \item \texttt{TableGPT} -- This approach from~\cite{li2023table} uses ``table fine-tuning'' to adapt large language models to a set of diverse table tasks, including entity matching. We report the F1 scores from \cite{li2023table} for the table-tuned model variants \texttt{TableGPT~[GPT-3.5-text-davinci-002]} and \texttt{TableGPT~[GPT-3.5-text-chat-davinci-002]} each of which have 175 billion parameters. We report the numbers from their zero-shot setting, where the task (entity matching) has been seen during training, but no labelled pairs for the target data have been observed. We can only report numbers for the full versions of seven out of our nine datasets though, and cannot compute numbers for the remaining two datasets (\texttt{ABT} and \texttt{WDC}) and the overall performance, since \texttt{TableGPT} is proprietary and not available to the academic community.
\end{itemize}

\header{Results and discussion} We list the resulting F1 scores for the nine datasets in \Cref{tab:zm_label}. Note that the best score per data is indicated in bold, while the second-best score is underlined. We report some of the scores for \texttt{Jellyfish} in brackets to indicate that it was pretrained on this data, which violates the zero-shot setting.

\headerl{Overall prediction quality} \texttt{MatchGPT} with the proprietary trillion-parameter model \texttt{GPT-4} achieves the highest average F1 score of 86.36. We find that \method{} provides the second-highest overall performance with a mean F1 score of 81.96. This is remarkable since it outperforms \texttt{MatchGPT} with all open and commercial models (except for GPT-4), which have up to three orders of magnitude more parameters than \method{}. We find that our method provides the best performance on the three datasets \texttt{BEER}, \texttt{FOZA} and \texttt{ITAM} from diverse domains and the second-best performance on \texttt{ABT} and \texttt{DBGO}.  However, for other datasets such as \texttt{AMGO}, \texttt{WAAM}, and \texttt{WDC}, there is still a  gap compared to \texttt{MatchGPT [GPT-4]}. We attribute this to the fact that these dataset employ very specific language to describe products, which is often not grammatically consistent. Our method for example misclassifies the matching product pair \texttt{\{name: "webroot spysweeper antispyware 3 user", price: "39.99"\}} and \texttt{\{name: "webroot software 65210 spy sweeper 3 pc", price: "31.99"\}} from the \texttt{AMGO} dataset, which \texttt{MatchGPT [GPT-4]} correctly matches. An unknown factor in these results is the question whether the commercial models have already seen the (publicly available) benchmark datasets at training time. This is impossible to determine however, since the training data of these models is not disclosed.

As expected, the \texttt{StringSim} baseline gives the worst average performance of 40.21. \texttt{TableGPT} (which again applies models with hundreds of billions of parameters) also scores high on the datasets for which its authors provide numbers in \cite{li2023table}. \method{} also outperforms \texttt{Jellyfish} (with a score of 81.35 compared to 77.83), even though the latter has seen data from six out the nine datasets at training time. The \texttt{Ditto} baseline (which we trained in a similar manner as our method and which has a similarly sized base model) also performs well with a score of 66.05 (and achieves on par performance with some of the larger models), but is more than 15 points behind our method, which validates the benefits of our model design. The parameterless \texttt{ZeroER} also scores high (and provides the best performance on \texttt{DBAC} and \texttt{FOZA}) and even outperforms \texttt{Ditto}. However, it still lags more than 14 points behind our method. 

In summary, we find that \method{} outperforms all baselines except for one and provides an average matching performance that is within 4.4\% percent of the best observed performance from  \texttt{MatchGPT [GPT-4]}, which requires a trillion parameter model (four orders of magnitude more parameters than \method{}).

\headerl{Trade-off between model size and prediction quality} An appealing property of \method{} is the comparatively small size of its base model GPT-2, which has 124 million parameters. In order to visualise this, we plot the average F1 score of several baseline models versus their number of parameters in \Cref{fig:f1-vs-size}. As discussed earlier \method{} drastically outperforms the similarly sized \texttt{Ditto} (which relies on the \texttt{Bert} model), and also provides a significantly higher mean F1 score than \texttt{MatchGPT} with \texttt{Mixtral}, \texttt{SOLAR}, \texttt{Beluga2}, \texttt{GPT-3.5-Turbo03} and \texttt{GPT-3.5-Turbo06}, which are roughly a 1,000 times larger (with parameter sizes ranging from 56 billion to 175 billion). This advantage in model size has major implications for the inference performance and cost, which we will evaluate in the following.


\begin{figure}[t!]
    \centering
    \begin{subfigure}{0.48\columnwidth}    
      \centering
      \includegraphics[width=\columnwidth]{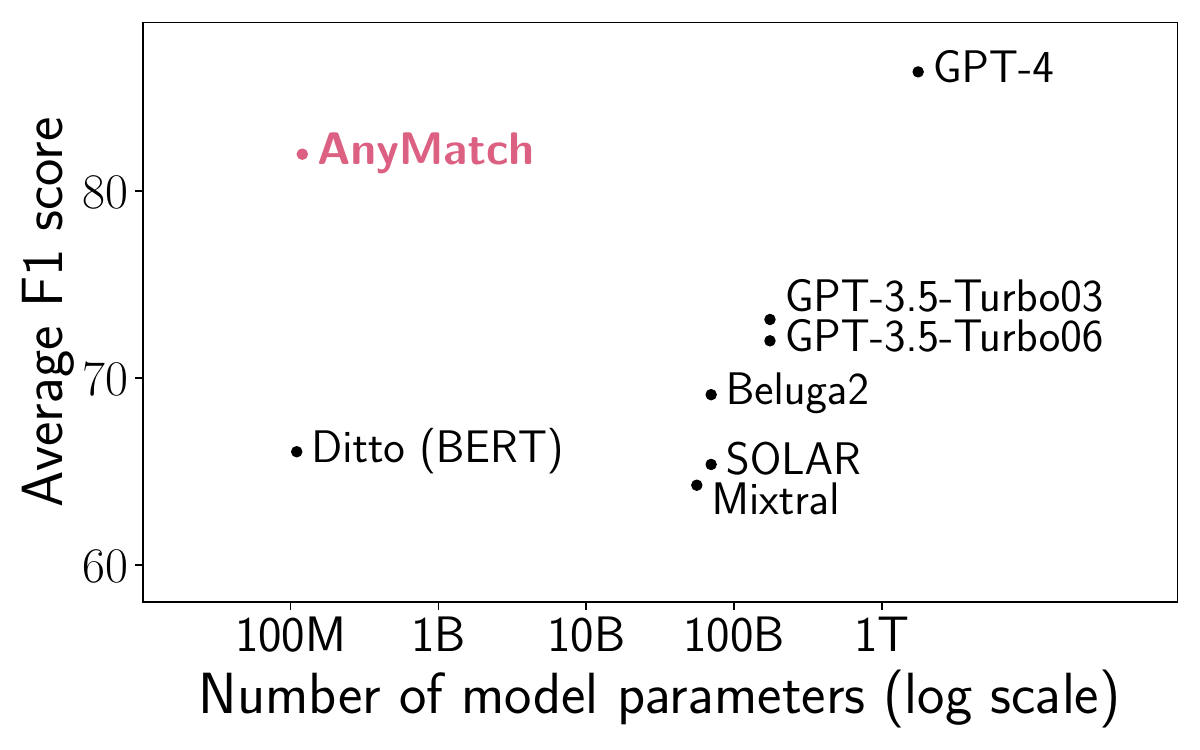}
      \caption{Model size versus prediction quality.}
      \label{fig:f1-vs-size}        
    \end{subfigure}      
    \hfill
    \begin{subfigure}{0.48\columnwidth}
      \centering
      \includegraphics[width=\columnwidth]{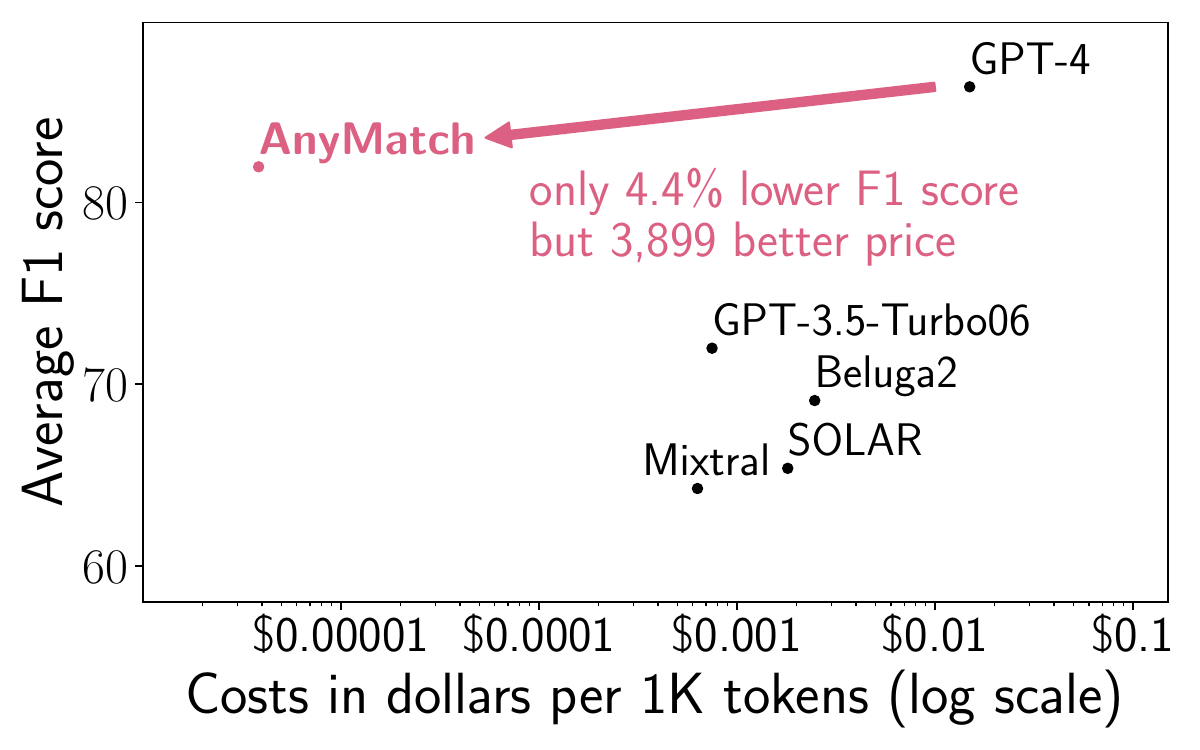}
      \caption{Deployment cost versus prediction quality.}
      \label{fig:f1-vs-cost}
    \end{subfigure}    
    \caption{\method{} outperforms the majority of models applied by \texttt{MatchGPT}, even though they have up to three orders of magnitude more parameters (a), and offers an attractive trade-off at a 3,899x better price than \texttt{MatchGPT} using \texttt{GPT-4} with only a 4.4\% decrease in prediction quality (b).}
    \label{fig:changes}
\end{figure}

\subsection{Inference Throughput \& Deployment~Cost}
\label{sec:eval-performance}

The goal of the next two experiments is to evaluate the computational performance and hardware requirements of \method{} in comparison to other zero-shot entity matching approaches. We focus on \texttt{MatchGPT} with GPT-4 (which outperforms \method{} by 4.4 percent in prediction quality) and \texttt{MatchGPT} with other LLMs, which also scored high in the evaluation. The ability to deploy zero-shot EM models in a cost-efficient and scalable way is especially important for the use cases discussed in \Cref{sec:problem}, such as data integration services in the cloud or deduplication as a data cleaning step in machine learning pipelines~\cite{xin2021production,li2021cleanml}. 
We focus on throughput as a measure of computational performance, as the input for entity matching are typically large candidate sets of potentially matching pairs, which have to be processed in batch. Since we focus on EM with language models, we adopt the common metric of tokens per second~\cite{borzunov2024distributed}. Unfortunately, we cannot properly evaluate the throughput of commercial models like GPT-4 since they are only accessible via proprietary APIs. It is unknown with which hardware they are run, and they are most likely also deployed for multi-tenancy, which makes performance evaluation even more difficult.

Therefore, we adopt the following methodology for our performance experiments: We measure and compare the throughput (in terms of tokens per second on a given hardware setup) of \method{} to the throughput of the matchers which leverage publicly available open-weight models (\Cref{sec:eval-performance-throughput}). Subsequently, we compare inference with the open-weight models to inference with proprietary model APIs in terms of cost rather than throughput. A common metric is the dollar price per 1,000 tokens, which is available for commercial models from OpenAI. We estimate this cost for \method{} and the open-weight models based on the observed throughput numbers and the hourly cost of an appropriate cloud instance in the Amazon Web Services Cloud~(\Cref{sec:eval-performance-cost}). Note that we cannot include the GPT-3.5 models used by \texttt{TableGPT} in these experiments, since they are already deprecated\footnote{\url{https://platform.openai.com/docs/deprecations}} and not available anymore. 

\subsubsection{Inference Throughput}
\label{sec:eval-performance-throughput}

The goal of our first experiment is to measure the inference throughput in terms of tokens per second of \method{} and the LLM-based methods \texttt{Jellyfish} (which uses the \texttt{Llama2-13B} model~\cite{touvron2023llama}) and \texttt{MatchGPT} with three open-weight models (\texttt{Mixtral-8x7B}, \texttt{SOLAR}, \texttt{Beluga2}).

\header{Experimental setup} We deploy each matcher (in combination with a given model) with exclusive access to a machine with four A100 GPUs with 40~GB GPU RAM in a large academic HPC cluster. Note that the A100 GPU is a common choice for ML, is the most powerful hardware available to us in academic context, and also constitutes a common choice in cloud instances designed for ML workloads. We leverage implementations based on PyTorch and the \texttt{transformers} library. We deploy quantised (16-bit precision) versions of the models and use model parallelism to distribute a model over multiple GPUs if it cannot fit into the 40~GB memory of a single A100 GPU.
We leverage the \texttt{DBGO} dataset here, since it is the largest dataset from our evaluation and proceed as follows. We first determine the maximum batch size usable per model by testing exponentially growing batch sizes and checking for memory issues. Next, we measure the inference time for 100 batches (based on the determined maximum batch size) via the \texttt{torch.utils.benchmark} package from PyTorch and compute the throughput in tokens/s based on the observed inference times. If a method does not use all four GPUs, we extrapolate its throughput to the full machine based on the number of GPUs used, as our inference is embarrassingly parallel.

\begin{table}[h!]
  \centering
  \resizebox{\linewidth}{!}{%
\begin{tabular}{l l r | r r r r }
\toprule
\textbf{Model} & \textbf{Used by} & \textbf{\#params} & \textbf{RAM}& \textbf{batch} & \textbf{Throughput} \\
 & & (millions) & (GB) & \textbf{size} & (tokens/s)\\
\midrule
\texttt{Llama2-13B} & \texttt{Jellyfish} & 13,000 & 24.46& 128 & \underline{26,721}\\
\texttt{Mixtral-8x7B} & \texttt{MatchGPT} & 56,000& 73.73& 32 & 2,108\\
\texttt{Beluga2} & \texttt{MatchGPT} & 70,000 & 128.64 & 32 & 1,079\\ 
\texttt{SOLAR} & \texttt{MatchGPT} & 70,000 & 128.64& 64 & 752\\
\midrule
\texttt{GPT-2} & \method{} & 124 & 0.26& 8,192 & \textbf{693,999} \\
\bottomrule
\end{tabular}}
  \caption{Throughput in tokens/s on a machine with 4xA100 (40GB) GPUs for different open-weight LLMs employed by \texttt{Jellyfish} and \texttt{MatchGPT} and our proposed model \method{}. Due to the small number of parameters in \method{}, its throughput is 25x higher than the throughput of \texttt{Jellyfish} and up to 922x higher than the throughput for the large models employed by \texttt{MatchGPT}. Note that \method{} outperforms these models in terms of prediction quality as well, despite its drastically smaller number of parameters.}
  \label{tab:throughput}  
\end{table}

\header{Results and discussion} We list the required memory per model, the corresponding maximum usable batch size and the achieved throughput in \Cref{tab:throughput}. Both \texttt{GPT-2} used by \method{} and \texttt{Llama2-13B} used by \texttt{Jellyfish} fit into the 40~GB memory of a single A100 GPU. \texttt{Mixtral-8x7B} requires model parallelism with two such GPUs, while \texttt{SOLAR} and \texttt{Beluga2} need to be distributed over all four A100 GPUs. The differences in size and the required model parallelism result in vast differences in the maximum achievable batch size and throughput. We observe that \method{} has an up to 922x times higher throughput than \texttt{MatchGPT} with its large models and a 25x times higher throughput than \texttt{Jellyfish}. We attribute the low throughput numbers for \texttt{MatchGPT} to the high memory requirements, which force the use of model parallelism over multiple GPUs. The latter is detrimental for throughput since the model activation must be copied over to the memory of other GPUs. \texttt{MatchGPT} can only use small batch sizes (16-32) and achieves a low throughput in the range of 752 to 2,108 tokens per second only. In contrast, the Llama2-13B model used by \texttt{Jellyfish} fits into the memory of a single GPU, works with a higher batch size of 128, and achieves a throughput of over 26 thousand tokens/s. Our proposed method \method{} can use a very large batch size of 8,192, since its base model GPT-2 has a comparably small parameter size of 124M, which only requires about 260M of GPU RAM for the model weights. This is two orders of magnitude less memory than \texttt{Jellyfish} and \texttt{MatchGPT}. As a result, \method{} achieves a throughput of over 690 thousand tokens/s in this experiment, which is 25x higher than the throughput of \method{} and between 329x to 922x higher than the throughput of \texttt{MatchGPT} with the large open-weight models. These throughput differences of up to two orders of magnitude are a clear indication of the performance benefits of leveraging a small language model for zero-shot entity matching.

\subsubsection{Inference Cost}
\label{sec:eval-performance-cost}

The goal of the following discussion is to compare \method{} against entity matching with approaches like \texttt{MatchGPT} that use commercial models from OpenAI. As discussed, we cannot reliably measure the throughput for these models since they are deployed behind proprietary APIs on unknown hardware. As a consequence, we compare the dollar cost of inference with these models to the dollar cost of inference with \method{} and \texttt{MatchGPT} with the open-weights models \texttt{Mixtral-8x7B}, \texttt{SOLAR} and \texttt{Beluga2}. 

\header{Setup} We lookup the costs for the commercial models from OpenAI at \url{https://openai.com/api/pricing/}. As of July 2024, batch inference with the \texttt{GPT-4} model costs \$0.015 per 1,000 tokens and inference with \texttt{GPT-3.5-turbo-0613} costs \$0.00075 per 1,000 tokens. Note that these models have different costs for input and output tokens; we use the cheaper input token cost, since entity matching is modelled as sequence classification task, which only generates a single output.

We estimate the cost for \method{} and the open-weight models as follows. We assume that such a model is deployed on a cloud instance that is constantly used for inference (e.g., as part of the use cases described in \Cref{sec:problem}). We use the cost for a \texttt{p4d.24xlarge} instance\footnote{\url{https://aws.amazon.com/ec2/instance-types/p4/}} from the Amazon Web Services cloud as a reference. This machine is designed for ML workloads and comes with eight A100 (40GB) GPUs (exactly twice the amount of GPUs which we used for our throughput experiment). As of June 2024, such a machine has an hourly cost of \$19.22 in a scenario where the instance is reserved for a year (which would be common in a corporate setup). Since the cloud instance has the exact same type of GPU (only twice the amount), we can extrapolate our throughput numbers from \Cref{sec:eval-performance-throughput} to this machine by simply doubling them, as inference in entity matching is an embarrassingly parallel workload. We therefore estimate the cost per 1,000 tokens for models deployed on this machine as $ (p / (2 \cdot t_m \cdot 3600)) \cdot 1000$ where $p$ is the hourly instance price, $t_m$ is the throughput in tokens/s observed for model $m$ and 2 is the extrapolation factor from our previous experiments (as the cloud instance has twice the amount of GPUs). 

For the open-weight models, we additionally lookup the hosting price on the cloud platform \texttt{together.ai}\footnote{Available at \url{https://www.together.ai/pricing}, accessed in June 2024} and choose this option if the resulting price per 1,000 tokens would be lower than our self-hosting setup (e.g., because a more favourable GPU can be chosen).

\begin{table}[h!]
  \centering
  \resizebox{\linewidth}{!}{%
\begin{tabular}{l l | l}
\toprule
 & \textbf{Cost for} & \\
\textbf{Method \& model} & \textbf{1K tokens} & \textbf{Deployment scenario}\\
\midrule
\texttt{MatchGPT [GPT-4]} & \$0.015 & OpenAI Batch API\\
\texttt{MatchGPT [SOLAR]} & \$0.0009 & Hosting on Together.ai\\
\texttt{MatchGPT [Beluga2]} & \$0.0009 & Hosting on Together.ai\\
\texttt{MatchGPT [GPT-3.5-turbo-06]} & \$0.00075 & OpenAI Batch API\\
\texttt{MatchGPT [Mixtral-8x7B]} & \$0.00063 & 4x on \texttt{p4d.24xlarge}\\
\texttt{Jellyfish} & \$0.000025 & 8x on \texttt{p4d.24xlarge}\\
\midrule
\method{} & \$0.0000038 & 8x on \texttt{p4d.24xlarge}\\
\bottomrule
\end{tabular}}
  \caption{Cost per 1K tokens for EM with proprietary models, compared to a deployment scenario with open-weight models on a \texttt{p4d.24xlarge} instance in the AWS cloud or via the together.ai platform. \method{} offers the lowest cost and three orders of magnitude cheaper than \texttt{MatchGPT} with the commercial \texttt{GPT-4} model.}
  \label{tab:cost}  
\end{table}

\header{Results and discussion} We list the resulting costs per method and model in descending order in \Cref{tab:cost} . For each entry, we also mention the chosen cheapest deployment scenario, e.g. whether we assume that the OpenAI API is used, whether we assume that the model is hosted on together.ai, or whether we assume that the model is deployed x-times on a \texttt{p4d.24xlarge} instance in AWS. We encounter the highest costs for \texttt{MatchGPT} with \texttt{GPT-4} (which gave the highest prediction quality). These costs are an order of magnitude higher than the cost for hosting \texttt{MatchGPT} with the \texttt{SOLAR} or \texttt{Beluga2} models or even using the older commercial model \texttt{GPT-3.5-turbo-06}. Using the smaller \texttt{Mixtral-8x7B} model is 23x cheaper than \texttt{GPT-4}, while \texttt{Jellyfish} (with a 13 billion parameter model) is already several orders of magnitude cheaper than \texttt{GPT-4}. \method{} is by far the cheapest model in this comparison, which is expected due to its low memory footprint and high throughout. It outperforms \texttt{GPT-4} by a factor of 3,899x in price. 

\header{Trade-off between deployment cost and prediction quality} We discuss the trade-off between the prediction quality of different approaches and their deployment cost. This trade-off is crucial for designing cost-efficient and scalable EM approaches, e.g. for a data integration service in the cloud. For that. we plot the average F1 score achieved versus the estimated cost for 1,000 tokens from the analysis in \Cref{fig:f1-vs-cost}. Note that we cannot include \texttt{TableGPT} in this discussion since we do not have scores from it for all datasets (and could not estimate its cost due to the deprecation of the used models). We also do not include \texttt{Jellyfish} in this discussion, since we cannot reliably compute its average F1 score, as it has seen several of the evaluation datasets at training time, which violates our zero-shot setting (as discussed in \Cref{sec:eval-quality}). 

\texttt{MatchGPT [GPT-4]} had the highest average F1-score of 86.36 for our nine datasets in a zero-shot setting. Our proposed model \method{} reached the second highest F1 score of 81.96, with only a 4.4\% difference, even though it has four orders of magnitude less parameters (124 million vs 1.76 trillion). The slight performance benefit of \texttt{MatchGPT [GPT-4]} comes at a drastic price increase though, as it has a 3,899 times higher inference cost than \method{}. These numbers showcase the attractive trade-off between prediction quality and deployment cost offered by \method{}, which suggest that \method{} should be preferred in scenarios where cost, scale and speed have priority over peak prediction quality. 

\subsection{Ablation Study}
\label{sec:eval-ablation}

\begin{table*}[t!] 
\centering
\resizebox{\linewidth}{!}{%
\begin{tabular}{l @{\hskip 0.3in} rrrrrrrrr @{\hskip 0.3in} l}
\toprule
 & \multicolumn{9}{c}{\textbf{Unseen Target Dataset}} &   \\
 & \texttt{ABT} & \texttt{AMGO} & \texttt{BEER} & \texttt{DBAC} & \texttt{DBGO} & \texttt{FOZA} & \texttt{ITAM} & \texttt{WAAM} & \texttt{WDC} & Mean \\
\midrule
\texttt{AnyMatch} (main model) & 86.05& 55.08& 96.55& 93.61& 90.59& 100.00& 90.91& 61.51& 63.31& 81.96\\
\midrule
\multicolumn{11}{l}{\textbf{Choice of base model} -- main model uses \texttt{GPT2}}\\
\midrule
(\texttt{GPT2}) $\rightarrow$ (\texttt{T5}) & 85.45& 56.12& 90.32& 87.49& 84.76& 97.67& 91.32& 60.47& 62.57& 79.57 ({\red $- \Delta 2.38$}) \\
(\texttt{GPT2}) $\rightarrow$ (\texttt{BERT}) & 73.82& 51.28& 87.46& 89.72& 86.54& 89.72& 72.65& 43.58& 61.42& 72.91 ({\red $- \Delta 9.04$})\\
\midrule
\multicolumn{11}{l}{\textbf{Design of serialisation format} -- main model uses \texttt{(suffix, <p> enclosement})}\\
\midrule
\texttt{(suffix, <p> enc.)} $\rightarrow$ \texttt{(suffix, <p> enc., column name)} & 84.02& 57.45& 90.32& 94.32& 88.33& 100.00& 94.55& 63.05& 61.79& 81.53 ({\red $- \Delta 0.43$})\\
\texttt{(suffix, <p> enc.)} $\rightarrow$ \texttt{(prefix, <p> enc.)} & 83.98& 55.12& 96.55& 92.64& 89.12& 97.67& 90.13& 60.12& 59.76& 80.56 ({\red $- \Delta 1.39$})\\
\texttt{(suffix, <p> enc.)} $\rightarrow$ \texttt{(prefix)} & 82.47& 54.27& 93.33& 92.76& 91.54& 100.00& 87.56& 59.83& 60.41& 80.24 ({\red $- \Delta 1.71$})\\
\midrule
\multicolumn{11}{l}{\textbf{Training data generation strategy} --- main model uses \texttt{(automl, flip, attr\_mix)}}\\
\midrule
\texttt{(automl, flip, attr\_mix)} $\rightarrow$ \texttt{(automl, attr\_mix)} & 84.01& 49.50& 96.55& 92.54& 90.12& 100.00& 89.79& 66.16& 62.35& 81.24 ({\red $- \Delta 0.72$})  \\
\texttt{(automl, flip, attr\_mix)} $\rightarrow$ \texttt{(attr\_mix)} & 82.27& 56.64& 90.32& 94.82& 89.98& 97.67& 87.72& 64.50& 59.11& 80.34 ({\red $- \Delta 1.62$}) \\
\texttt{(automl, flip, attr\_mix)} $\rightarrow$ \texttt{(attr\_seq)} & 84.03& 58.65& 84.85& 95.18& 90.26& 91.67& 88.89& 57.05& 58.77& 78.82  ({\red $- \Delta 3.14$}) \\
\texttt{(automl, flip, attr\_mix)} $\rightarrow$ \texttt{()} & 82.82& 52.91& 87.50& 93.79& 89.58& 97.67& 90.91& 55.71& 58.05& 78.77  ({\red $- \Delta 3.19$}) \\
\bottomrule
\end{tabular}}
\addtolength{\tabcolsep}{1pt}
\caption{F1 scores from our ablation study for the choice of base model, design of serialisation format and the training data generation strategies of \method{}. Altering any of our design choices leads to a performance decrease of up to 3.19 percent.}
\label{tab:ablation}
\end{table*}

Finally, we conduct an ablation study for \method{} in order to validate our design decisions from \Cref{sec:approach}. For that, we remove and/or replace different components of our model and show that this  removal and replacement results in a performance decrease. We evaluate the resulting model variants analogously to \Cref{sec:eval-quality}, where we measure the average F1 score over the nine benchmark datasets in a zero-shot setting. We summarise the tested variants together with the corresponding results for the individual datasets in \Cref{tab:ablation}. In addition, we report the performance delta (the reduction in average F1 score) for all tested variants, in comparison to the proposed design of \method{}.

\subsubsection{Choice of base model} The goal of our first experiment is to validate the choice of the decoder-only language model \texttt{GPT-2}~\cite{radford2019language} as base model for \method{}.

\header{Experimental setup} We evaluate the impact of replacing \texttt{GPT-2} in \method{} with different similarly sized alternative models. In particular, we evaluate Google's \texttt{T5}~\cite{raffel2020exploring} model with an encoder-decoder architecture and the encoder-only model \texttt{Bert}~\cite{devlin2018bert}, which is for example also used by \cite{li2020deep}. 

\header{Results and discussion} We find that both alternative models result in a decrease of the overall F1 score, as detailed in \Cref{tab:ablation}: using \texttt{T5} results in a decrease of 2.38\%, while using \texttt{Bert} leads to the drastic decrease of 9.04\% in average F1 score. The performance loss observed when switching to the \texttt{T5} model likely stems from its better suitability for sequence-to-sequence tasks \cite{raffel2020exploring}, whereas we treat our task as sequence classification and have specifically tailored the prompt design for this purpose. 
Moreover, we attribute the significant decrease in predictive quality when using \texttt{Bert} to the following factors: \texttt{Bert} encodes the input into a vectorised representation, to which a prediction head is subsequently appended for making predictions. However, in our approach, we incorporate a task description into the input sequence, accounting for roughly 5\% of the total input, which potentially negatively influences the downstream classification. Note while using a different serialization method for the \texttt{Bert} model might improve performance, this is not the focus of this ablation study.

\subsubsection{Choice of serialization format} The goal of this experiment is to validate the design of our serialization format. In particular, we enclose the serialized record pairs in \texttt{<p>} tags (referred to as  \texttt{<p> enclosement}), use the placeholder \texttt{COL} to denote the starting position of each attribute value, and add the question \textit{``Given the attributes of the two records, are they the same?''} as a \texttt{suffix} to the end of the model prompt.

\header{Experimental setup} We evaluate three alternative serialization formats. The first variant \texttt{(prefix, <p> enclosement, column name}) replaces the COL placeholder with the actual column name to specify attributes.
In the second variant called  \texttt{(prefix, <p> enclosement}), we prepend the question to the prompt. In the last variant  \texttt{(prefix}), we additionally remove the \texttt{<p> enclosement} from the format.

\header{Results and discussion} As detailed in \Cref{tab:ablation}, all variants lead to a loss in prediction quality. Including the column name lead to a 0.43\% performance loss. This is an indication that the reliance on attribute names at training time undermines the model's transferability as the target data lacks such information. Changing the \texttt{suffix} format to \texttt{prefix} results in a drop of 1.39\% in terms of average F1, while additionally removing the \texttt{<p> enclosement} decreases the drop to~1.71\%. 

\subsubsection{Benefits of difficult pairs selection and attribute-level matching during training data generation} Next, we aim to validate the choice of the different techniques used during the generation of training data for \method{}. 

\header{Experimental setup} We show that these techniques contribute to the predictive performance of \method{}. For that we implement different variants of our training data generation algorithm and evaluate the zero-shot performance of our model trained on this data on the nine datasets analogous to the setting in \Cref{sec:eval-quality}. As discussed, our main model uses difficult pair selection via AutoML (\texttt{automl}) for record-level instances, augments the data with ``flipped'' record pairs (\texttt{flip}) and mixes attribute-level training instances into the training data (\texttt{attr\_mix}). In this experiment, we evaluate the following reduced variants:
\begin{itemize}[leftmargin=*]
  \item \texttt{(automl, attr\_mix)} -- This variant does not include flipped record pairs.
  \item \texttt{(attr\_mix)} -- This variant does not use our AutoML-based approach to select difficult matching examples and does not include flipped record pairs.
  \item \texttt{(attr\_seq)} -- This variant also does not use our AutoML-based approach to select difficult matching examples and does not include flipped record pairs. Additionally, it does not mix the attribute-level training instances with the record-level training instances. Instead, it uses sequential training to fine-tune the model on the attribute-level pairs before continuing the fine-tuning on record-level examples.
  \item \texttt{()} -- This variant neither uses the AutoML-based approach to select difficult matching examples nor any attribute-level training examples or flipped record pairs.
\end{itemize}

\header{Results and discussion} We list the resulting F1 scores (and the deltas in F1 compared to the main model) in \Cref{tab:ablation}. The results validate the benefits of our training data generation techniques, as we find that the removal of any of our proposed techniques results in a performance decrease. Removing the flipped records pairs in \texttt{(automl, attr\_mix)} leads to a decrease of 0.72\%, additionally removing the selection of difficult examples via AutoML in \texttt{(attr\_mix)} further decreases the performance by 1.62\%. Removing  or not mixing in the attribute level instances (e.g., the variants \texttt{(attr\_seq)} and \texttt{()}) leads to a performance loss of more than 3\%.

\vspace{2mm}
In summary, the ablation study confirms that all our design decisions for the model and fine-tuning data generation contribute to the predictive performance of \method{}.

\section{Conclusion}
We revisited the zero-shot EM problem with \method{}, a small language model fine-tuned in a transfer learning setup, and proposed several novel data selection techniques to generate high-quality fine-tuning data for our model. 

We conducted an extensive evaluation of the prediction quality and deployment cost of our model, in a comparison to thirteen baselines on nine benchmark datasets. We find that \method{} provides competitive prediction quality despite its small parameter size and exhibits drastic deployment cost benefits compared to the latest EM approaches, which leverage trillion parameter LLMs.

\header{Future work} In future work, we aim to investigate whether we can combine \method{} with expensive LLMs like \texttt{GPT-4} in a smart way that still retains the overall cost-efficiency, e.g., by using \texttt{GPT-4} only for samples where our method has low confidence. Furthermore, we also plan to extend \method{} to make use of additional labelled samples, e.g., in few-shot scenarios.
\balance

\bibliographystyle{ACM-Reference-Format}
\bibliography{zeromatch}


\begin{thebibliography}{43}


\ifx \showCODEN    \undefined \def \showCODEN     #1{\unskip}     \fi
\ifx \showDOI      \undefined \def \showDOI       #1{#1}\fi
\ifx \showISBNx    \undefined \def \showISBNx     #1{\unskip}     \fi
\ifx \showISBNxiii \undefined \def \showISBNxiii  #1{\unskip}     \fi
\ifx \showISSN     \undefined \def \showISSN      #1{\unskip}     \fi
\ifx \showLCCN     \undefined \def \showLCCN      #1{\unskip}     \fi
\ifx \shownote     \undefined \def \shownote      #1{#1}          \fi
\ifx \showarticletitle \undefined \def \showarticletitle #1{#1}   \fi
\ifx \showURL      \undefined \def \showURL       {\relax}        \fi
\providecommand\bibfield[2]{#2}
\providecommand\bibinfo[2]{#2}
\providecommand\natexlab[1]{#1}
\providecommand\showeprint[2][]{arXiv:#2}

\bibitem[Abdelaal et~al\mbox{.}(2023)]%
        {rein2022}
\bibfield{author}{\bibinfo{person}{Mohamed Abdelaal}, \bibinfo{person}{Christian Hammacher}, {and} \bibinfo{person}{Harald Schoening}.} \bibinfo{year}{2023}\natexlab{}.
\newblock \showarticletitle{REIN: A Comprehensive Benchmark Framework for Data Cleaning Methods in ML Pipelines}.
\newblock \bibinfo{journal}{\emph{Proceedings of the VLDB Endowment (PVLDB)}} (\bibinfo{year}{2023}).
\newblock


\bibitem[Abedjan et~al\mbox{.}(2016)]%
        {abedjan2016detecting}
\bibfield{author}{\bibinfo{person}{Ziawasch Abedjan}, \bibinfo{person}{Xu Chu}, \bibinfo{person}{Dong Deng}, \bibinfo{person}{Raul~Castro Fernandez}, \bibinfo{person}{Ihab~F Ilyas}, \bibinfo{person}{Mourad Ouzzani}, \bibinfo{person}{Paolo Papotti}, \bibinfo{person}{Michael Stonebraker}, {and} \bibinfo{person}{Nan Tang}.} \bibinfo{year}{2016}\natexlab{}.
\newblock \showarticletitle{Detecting data errors: Where are we and what needs to be done?}
\newblock \bibinfo{journal}{\emph{Proceedings of the VLDB Endowment}} \bibinfo{volume}{9}, \bibinfo{number}{12} (\bibinfo{year}{2016}), \bibinfo{pages}{993--1004}.
\newblock


\bibitem[Badaro et~al\mbox{.}(2023)]%
        {badaro2023transformers}
\bibfield{author}{\bibinfo{person}{Gilbert Badaro}, \bibinfo{person}{Mohammed Saeed}, {and} \bibinfo{person}{Paolo Papotti}.} \bibinfo{year}{2023}\natexlab{}.
\newblock \showarticletitle{Transformers for tabular data representation: A survey of models and applications}.
\newblock \bibinfo{journal}{\emph{Transactions of the Association for Computational Linguistics}}  \bibinfo{volume}{11} (\bibinfo{year}{2023}), \bibinfo{pages}{227--249}.
\newblock


\bibitem[Borzunov et~al\mbox{.}(2024)]%
        {borzunov2024distributed}
\bibfield{author}{\bibinfo{person}{Alexander Borzunov}, \bibinfo{person}{Max Ryabinin}, \bibinfo{person}{Artem Chumachenko}, \bibinfo{person}{Dmitry Baranchuk}, \bibinfo{person}{Tim Dettmers}, \bibinfo{person}{Younes Belkada}, \bibinfo{person}{Pavel Samygin}, {and} \bibinfo{person}{Colin~A Raffel}.} \bibinfo{year}{2024}\natexlab{}.
\newblock \showarticletitle{Distributed inference and fine-tuning of large language models over the internet}.
\newblock \bibinfo{journal}{\emph{Advances in Neural Information Processing Systems}}  \bibinfo{volume}{36} (\bibinfo{year}{2024}).
\newblock


\bibitem[Brown et~al\mbox{.}(2020)]%
        {brown2020language}
\bibfield{author}{\bibinfo{person}{Tom Brown}, \bibinfo{person}{Benjamin Mann}, \bibinfo{person}{Nick Ryder}, \bibinfo{person}{Melanie Subbiah}, \bibinfo{person}{Jared~D Kaplan}, \bibinfo{person}{Prafulla Dhariwal}, \bibinfo{person}{Arvind Neelakantan}, \bibinfo{person}{Pranav Shyam}, \bibinfo{person}{Girish Sastry}, \bibinfo{person}{Amanda Askell}, {et~al\mbox{.}}} \bibinfo{year}{2020}\natexlab{}.
\newblock \showarticletitle{Language models are few-shot learners}.
\newblock \bibinfo{journal}{\emph{Advances in neural information processing systems}}  \bibinfo{volume}{33} (\bibinfo{year}{2020}), \bibinfo{pages}{1877--1901}.
\newblock


\bibitem[Chen et~al\mbox{.}(2021)]%
        {chen2021gnem}
\bibfield{author}{\bibinfo{person}{Runjin Chen}, \bibinfo{person}{Yanyan Shen}, {and} \bibinfo{person}{Dongxiang Zhang}.} \bibinfo{year}{2021}\natexlab{}.
\newblock \showarticletitle{GNEM: a generic one-to-set neural entity matching framework}. In \bibinfo{booktitle}{\emph{Proceedings of the Web Conference 2021}}. \bibinfo{pages}{1686--1694}.
\newblock


\bibitem[Devlin et~al\mbox{.}(2018)]%
        {devlin2018bert}
\bibfield{author}{\bibinfo{person}{Jacob Devlin}, \bibinfo{person}{Ming-Wei Chang}, \bibinfo{person}{Kenton Lee}, {and} \bibinfo{person}{Kristina Toutanova}.} \bibinfo{year}{2018}\natexlab{}.
\newblock \showarticletitle{Bert: Pre-training of deep bidirectional transformers for language understanding}.
\newblock \bibinfo{journal}{\emph{arXiv preprint arXiv:1810.04805}} (\bibinfo{year}{2018}).
\newblock


\bibitem[Doan et~al\mbox{.}(2020)]%
        {doan2020magellan}
\bibfield{author}{\bibinfo{person}{AnHai Doan}, \bibinfo{person}{Pradap Konda}, \bibinfo{person}{Paul Suganthan~GC}, \bibinfo{person}{Yash Govind}, \bibinfo{person}{Derek Paulsen}, \bibinfo{person}{Kaushik Chandrasekhar}, \bibinfo{person}{Philip Martinkus}, {and} \bibinfo{person}{Matthew Christie}.} \bibinfo{year}{2020}\natexlab{}.
\newblock \showarticletitle{Magellan: toward building ecosystems of entity matching solutions}.
\newblock \bibinfo{journal}{\emph{Commun. ACM}} \bibinfo{volume}{63}, \bibinfo{number}{8} (\bibinfo{year}{2020}), \bibinfo{pages}{83--91}.
\newblock


\bibitem[Erickson et~al\mbox{.}(2020)]%
        {erickson2020autogluon}
\bibfield{author}{\bibinfo{person}{Nick Erickson}, \bibinfo{person}{Jonas Mueller}, \bibinfo{person}{Alexander Shirkov}, \bibinfo{person}{Hang Zhang}, \bibinfo{person}{Pedro Larroy}, \bibinfo{person}{Mu Li}, {and} \bibinfo{person}{Alexander Smola}.} \bibinfo{year}{2020}\natexlab{}.
\newblock \showarticletitle{Autogluon-tabular: Robust and accurate automl for structured data}.
\newblock \bibinfo{journal}{\emph{arXiv preprint arXiv:2003.06505}} (\bibinfo{year}{2020}).
\newblock


\bibitem[Fan et~al\mbox{.}(2024)]%
        {fan2024gen}
\bibfield{author}{\bibinfo{person}{Grace Fan}, \bibinfo{person}{Roee Shraga}, {and} \bibinfo{person}{Ren{\'e}e~J Miller}.} \bibinfo{year}{2024}\natexlab{}.
\newblock \showarticletitle{Gen-T: Table Reclamation in Data Lakes}.
\newblock \bibinfo{journal}{\emph{arXiv preprint arXiv:2403.14128}} (\bibinfo{year}{2024}).
\newblock


\bibitem[Fernandez et~al\mbox{.}(2023)]%
        {fernandez2023large}
\bibfield{author}{\bibinfo{person}{Raul~Castro Fernandez}, \bibinfo{person}{Aaron~J Elmore}, \bibinfo{person}{Michael~J Franklin}, \bibinfo{person}{Sanjay Krishnan}, {and} \bibinfo{person}{Chenhao Tan}.} \bibinfo{year}{2023}\natexlab{}.
\newblock \showarticletitle{How large language models will disrupt data management}.
\newblock \bibinfo{journal}{\emph{Proceedings of the VLDB Endowment}} \bibinfo{volume}{16}, \bibinfo{number}{11} (\bibinfo{year}{2023}), \bibinfo{pages}{3302--3309}.
\newblock


\bibitem[Fu et~al\mbox{.}(2021)]%
        {fu2021hierarchical}
\bibfield{author}{\bibinfo{person}{Cheng Fu}, \bibinfo{person}{Xianpei Han}, \bibinfo{person}{Jiaming He}, {and} \bibinfo{person}{Le Sun}.} \bibinfo{year}{2021}\natexlab{}.
\newblock \showarticletitle{Hierarchical matching network for heterogeneous entity resolution}. In \bibinfo{booktitle}{\emph{Proceedings of the Twenty-Ninth International Conference on International Joint Conferences on Artificial Intelligence}}. \bibinfo{pages}{3665--3671}.
\newblock


\bibitem[Gao et~al\mbox{.}(2018)]%
        {gao2018navigating}
\bibfield{author}{\bibinfo{person}{Yihan Gao}, \bibinfo{person}{Silu Huang}, {and} \bibinfo{person}{Aditya Parameswaran}.} \bibinfo{year}{2018}\natexlab{}.
\newblock \showarticletitle{Navigating the data lake with datamaran: Automatically extracting structure from log datasets}. In \bibinfo{booktitle}{\emph{Proceedings of the 2018 International Conference on Management of Data}}. \bibinfo{pages}{943--958}.
\newblock


\bibitem[Huang and Wu(2024)]%
        {huang2024relationalizing}
\bibfield{author}{\bibinfo{person}{Zezhou Huang} {and} \bibinfo{person}{Eugene Wu}.} \bibinfo{year}{2024}\natexlab{}.
\newblock \showarticletitle{Relationalizing Tables with Large Language Models: The Promise and Challenges}. In \bibinfo{booktitle}{\emph{2024 IEEE 40th International Conference on Data Engineering Workshops (ICDEW)}}. IEEE, \bibinfo{pages}{305--309}.
\newblock


\bibitem[Hynes et~al\mbox{.}(2017)]%
        {hyneslinter}
\bibfield{author}{\bibinfo{person}{Nick Hynes}, \bibinfo{person}{D. Sculley}, {and} \bibinfo{person}{Michael Terry}.} \bibinfo{year}{2017}\natexlab{}.
\newblock \showarticletitle{The Data Linter: Lightweight Automated Sanity Checking for ML Data Sets}.
\newblock \bibinfo{journal}{\emph{Machine Learning Systems workshop at NeurIPS}} (\bibinfo{year}{2017}).
\newblock


\bibitem[Leone et~al\mbox{.}(2022)]%
        {leone2022critical}
\bibfield{author}{\bibinfo{person}{Manuel Leone}, \bibinfo{person}{Stefano Huber}, \bibinfo{person}{Akhil Arora}, \bibinfo{person}{Alberto Garc{\'\i}a-Dur{\'a}n}, {and} \bibinfo{person}{Robert West}.} \bibinfo{year}{2022}\natexlab{}.
\newblock \showarticletitle{A critical re-evaluation of neural methods for entity alignment}.
\newblock \bibinfo{journal}{\emph{Proceedings of the VLDB Endowment}} \bibinfo{volume}{15}, \bibinfo{number}{8} (\bibinfo{year}{2022}), \bibinfo{pages}{1712--1725}.
\newblock


\bibitem[Li et~al\mbox{.}(2024)]%
        {li2023table}
\bibfield{author}{\bibinfo{person}{Peng Li}, \bibinfo{person}{Yeye He}, \bibinfo{person}{Dror Yashar}, \bibinfo{person}{Weiwei Cui}, \bibinfo{person}{Song Ge}, \bibinfo{person}{Haidong Zhang}, \bibinfo{person}{Danielle Rifinski~Fainman}, \bibinfo{person}{Dongmei Zhang}, {and} \bibinfo{person}{Surajit Chaudhuri}.} \bibinfo{year}{2024}\natexlab{}.
\newblock \showarticletitle{Table-GPT: Table Fine-tuned GPT for Diverse Table Tasks}.
\newblock \bibinfo{journal}{\emph{Proceedings of the ACM on Management of Data}} \bibinfo{volume}{2}, \bibinfo{number}{3} (\bibinfo{year}{2024}), \bibinfo{pages}{1--28}.
\newblock


\bibitem[Li et~al\mbox{.}(2021)]%
        {li2021cleanml}
\bibfield{author}{\bibinfo{person}{Peng Li}, \bibinfo{person}{Xi Rao}, \bibinfo{person}{Jennifer Blase}, \bibinfo{person}{Yue Zhang}, \bibinfo{person}{Xu Chu}, {and} \bibinfo{person}{Ce Zhang}.} \bibinfo{year}{2021}\natexlab{}.
\newblock \showarticletitle{CleanML: A study for evaluating the impact of data cleaning on ml classification tasks}. In \bibinfo{booktitle}{\emph{2021 IEEE 37th International Conference on Data Engineering (ICDE)}}. IEEE, \bibinfo{pages}{13--24}.
\newblock


\bibitem[Li et~al\mbox{.}(2020)]%
        {li2020deep}
\bibfield{author}{\bibinfo{person}{Yuliang Li}, \bibinfo{person}{Jinfeng Li}, \bibinfo{person}{Yoshihiko Suhara}, \bibinfo{person}{AnHai Doan}, {and} \bibinfo{person}{Wang-Chiew Tan}.} \bibinfo{year}{2020}\natexlab{}.
\newblock \showarticletitle{Deep entity matching with pre-trained language models}.
\newblock \bibinfo{journal}{\emph{arXiv preprint arXiv:2004.00584}} (\bibinfo{year}{2020}).
\newblock


\bibitem[Liu et~al\mbox{.}(2024)]%
        {liu2024declarative}
\bibfield{author}{\bibinfo{person}{Chunwei Liu}, \bibinfo{person}{Matthew Russo}, \bibinfo{person}{Michael Cafarella}, \bibinfo{person}{Lei Cao}, \bibinfo{person}{Peter~Baille Chen}, \bibinfo{person}{Zui Chen}, \bibinfo{person}{Michael Franklin}, \bibinfo{person}{Tim Kraska}, \bibinfo{person}{Samuel Madden}, {and} \bibinfo{person}{Gerardo Vitagliano}.} \bibinfo{year}{2024}\natexlab{}.
\newblock \showarticletitle{A Declarative System for Optimizing AI Workloads}.
\newblock \bibinfo{journal}{\emph{arXiv preprint arXiv:2405.14696}} (\bibinfo{year}{2024}).
\newblock


\bibitem[Mudgal et~al\mbox{.}(2018)]%
        {mudgal2018deep}
\bibfield{author}{\bibinfo{person}{Sidharth Mudgal}, \bibinfo{person}{Han Li}, \bibinfo{person}{Theodoros Rekatsinas}, \bibinfo{person}{AnHai Doan}, \bibinfo{person}{Youngchoon Park}, \bibinfo{person}{Ganesh Krishnan}, \bibinfo{person}{Rohit Deep}, \bibinfo{person}{Esteban Arcaute}, {and} \bibinfo{person}{Vijay Raghavendra}.} \bibinfo{year}{2018}\natexlab{}.
\newblock \showarticletitle{Deep learning for entity matching: A design space exploration}. In \bibinfo{booktitle}{\emph{Proceedings of the 2018 international conference on management of data}}. \bibinfo{pages}{19--34}.
\newblock


\bibitem[Narayan et~al\mbox{.}(2022)]%
        {narayanfm2022}
\bibfield{author}{\bibinfo{person}{Avanika Narayan} {et~al\mbox{.}}} \bibinfo{year}{2022}\natexlab{}.
\newblock \showarticletitle{Can Foundation Models Wrangle Your Data?}
\newblock \bibinfo{journal}{\emph{PVLDB}} (\bibinfo{year}{2022}).
\newblock


\bibitem[Papadakis et~al\mbox{.}(2024)]%
        {papadakis2023critical}
\bibfield{author}{\bibinfo{person}{George Papadakis}, \bibinfo{person}{Nishadi Kirielle}, \bibinfo{person}{Peter Christen}, {and} \bibinfo{person}{Themis Palpanas}.} \bibinfo{year}{2024}\natexlab{}.
\newblock \showarticletitle{A critical re-evaluation of benchmark datasets for (deep) learning-based matching algorithms}.
\newblock \bibinfo{journal}{\emph{ICDE}} (\bibinfo{year}{2024}).
\newblock


\bibitem[Papadakis et~al\mbox{.}(2020)]%
        {papadakis2020blocking}
\bibfield{author}{\bibinfo{person}{George Papadakis}, \bibinfo{person}{Dimitrios Skoutas}, \bibinfo{person}{Emmanouil Thanos}, {and} \bibinfo{person}{Themis Palpanas}.} \bibinfo{year}{2020}\natexlab{}.
\newblock \showarticletitle{Blocking and filtering techniques for entity resolution: A survey}.
\newblock \bibinfo{journal}{\emph{ACM Computing Surveys (CSUR)}} \bibinfo{volume}{53}, \bibinfo{number}{2} (\bibinfo{year}{2020}), \bibinfo{pages}{1--42}.
\newblock


\bibitem[Paszke et~al\mbox{.}(2019)]%
        {paszke2019pytorch}
\bibfield{author}{\bibinfo{person}{Adam Paszke}, \bibinfo{person}{Sam Gross}, \bibinfo{person}{Francisco Massa}, \bibinfo{person}{Adam Lerer}, \bibinfo{person}{James Bradbury}, \bibinfo{person}{Gregory Chanan}, \bibinfo{person}{Trevor Killeen}, \bibinfo{person}{Zeming Lin}, \bibinfo{person}{Natalia Gimelshein}, \bibinfo{person}{Luca Antiga}, {et~al\mbox{.}}} \bibinfo{year}{2019}\natexlab{}.
\newblock \showarticletitle{Pytorch: An imperative style, high-performance deep learning library}.
\newblock \bibinfo{journal}{\emph{Advances in neural information processing systems}}  \bibinfo{volume}{32} (\bibinfo{year}{2019}).
\newblock


\bibitem[Peeters and Bizer(2023)]%
        {peeters2023entity}
\bibfield{author}{\bibinfo{person}{Ralph Peeters} {and} \bibinfo{person}{Christian Bizer}.} \bibinfo{year}{2023}\natexlab{}.
\newblock \showarticletitle{Entity matching using large language models}.
\newblock \bibinfo{journal}{\emph{arXiv preprint arXiv:2310.11244}} (\bibinfo{year}{2023}).
\newblock


\bibitem[Radford et~al\mbox{.}(2019)]%
        {radford2019language}
\bibfield{author}{\bibinfo{person}{Alec Radford}, \bibinfo{person}{Jeffrey Wu}, \bibinfo{person}{Rewon Child}, \bibinfo{person}{David Luan}, \bibinfo{person}{Dario Amodei}, {and} \bibinfo{person}{Ilya Sutskever}.} \bibinfo{year}{2019}\natexlab{}.
\newblock \showarticletitle{Language models are unsupervised multitask learners}.
\newblock \bibinfo{journal}{\emph{OpenAI Blog}} \bibinfo{volume}{1}, \bibinfo{number}{8} (\bibinfo{year}{2019}), \bibinfo{pages}{9}.
\newblock


\bibitem[Raffel et~al\mbox{.}(2020)]%
        {raffel2020exploring}
\bibfield{author}{\bibinfo{person}{Colin Raffel}, \bibinfo{person}{Noam Shazeer}, \bibinfo{person}{Adam Roberts}, \bibinfo{person}{Katherine Lee}, \bibinfo{person}{Sharan Narang}, \bibinfo{person}{Michael Matena}, \bibinfo{person}{Yanqi Zhou}, \bibinfo{person}{Wei Li}, {and} \bibinfo{person}{Peter~J Liu}.} \bibinfo{year}{2020}\natexlab{}.
\newblock \showarticletitle{Exploring the limits of transfer learning with a unified text-to-text transformer}.
\newblock \bibinfo{journal}{\emph{Journal of machine learning research}} \bibinfo{volume}{21}, \bibinfo{number}{140} (\bibinfo{year}{2020}), \bibinfo{pages}{1--67}.
\newblock


\bibitem[Services(2022a)]%
        {glue}
\bibfield{author}{\bibinfo{person}{Amazon~Web Services}.} \bibinfo{year}{2022}\natexlab{a}.
\newblock \bibinfo{title}{{AWS Glue}}.
\newblock
\newblock
\newblock
\shownote{\url{https://aws.amazon.com/glue/}}.


\bibitem[Services(2022b)]%
        {glueteaching}
\bibfield{author}{\bibinfo{person}{Amazon~Web Services}.} \bibinfo{year}{2022}\natexlab{b}.
\newblock \bibinfo{title}{Teaching the Find Matches transform}.
\newblock
\newblock
\newblock
\shownote{\url{https://docs.aws.amazon.com/glue/latest/dg/machine-learning-teaching.html}}.


\bibitem[Shahbazi et~al\mbox{.}(2023)]%
        {shahbazi2023through}
\bibfield{author}{\bibinfo{person}{Nima Shahbazi}, \bibinfo{person}{Nikola Danevski}, \bibinfo{person}{Fatemeh Nargesian}, \bibinfo{person}{Abolfazl Asudeh}, {and} \bibinfo{person}{Divesh Srivastava}.} \bibinfo{year}{2023}\natexlab{}.
\newblock \showarticletitle{Through the Fairness Lens: Experimental Analysis and Evaluation of Entity Matching}.
\newblock \bibinfo{journal}{\emph{Proceedings of the VLDB Endowment}} \bibinfo{volume}{16}, \bibinfo{number}{11} (\bibinfo{year}{2023}), \bibinfo{pages}{3279--3292}.
\newblock


\bibitem[Stonebraker et~al\mbox{.}(2018)]%
        {stonebraker2018data}
\bibfield{author}{\bibinfo{person}{Michael Stonebraker}, \bibinfo{person}{Ihab~F Ilyas}, {et~al\mbox{.}}} \bibinfo{year}{2018}\natexlab{}.
\newblock \showarticletitle{Data Integration: The Current Status and the Way Forward.}
\newblock \bibinfo{journal}{\emph{IEEE Data Eng. Bull.}} \bibinfo{volume}{41}, \bibinfo{number}{2} (\bibinfo{year}{2018}), \bibinfo{pages}{3--9}.
\newblock


\bibitem[Touvron et~al\mbox{.}(2023)]%
        {touvron2023llama}
\bibfield{author}{\bibinfo{person}{Hugo Touvron}, \bibinfo{person}{Louis Martin}, \bibinfo{person}{Kevin Stone}, \bibinfo{person}{Peter Albert}, \bibinfo{person}{Amjad Almahairi}, \bibinfo{person}{Yasmine Babaei}, \bibinfo{person}{Nikolay Bashlykov}, \bibinfo{person}{Soumya Batra}, \bibinfo{person}{Prajjwal Bhargava}, \bibinfo{person}{Shruti Bhosale}, {et~al\mbox{.}}} \bibinfo{year}{2023}\natexlab{}.
\newblock \showarticletitle{Llama 2: Open foundation and fine-tuned chat models}.
\newblock \bibinfo{journal}{\emph{arXiv preprint arXiv:2307.09288}} (\bibinfo{year}{2023}).
\newblock


\bibitem[Vos et~al\mbox{.}(2022)]%
        {vos2022towards}
\bibfield{author}{\bibinfo{person}{David Vos}, \bibinfo{person}{Till D{\"o}hmen}, {and} \bibinfo{person}{Sebastian Schelter}.} \bibinfo{year}{2022}\natexlab{}.
\newblock \showarticletitle{Towards parameter-efficient automation of data wrangling tasks with prefix-tuning}. In \bibinfo{booktitle}{\emph{NeurIPS 2022 First Table Representation Workshop}}.
\newblock


\bibitem[Wang et~al\mbox{.}(2022)]%
        {wang2022machop}
\bibfield{author}{\bibinfo{person}{Jin Wang}, \bibinfo{person}{Yuliang Li}, \bibinfo{person}{Wataru Hirota}, {and} \bibinfo{person}{Eser Kandogan}.} \bibinfo{year}{2022}\natexlab{}.
\newblock \showarticletitle{Machop: an end-to-end generalized entity matching framework}. In \bibinfo{booktitle}{\emph{Proceedings of the Fifth International Workshop on Exploiting Artificial Intelligence Techniques for Data Management}}. \bibinfo{pages}{1--10}.
\newblock


\bibitem[Wang et~al\mbox{.}(2024)]%
        {wang2024mixture}
\bibfield{author}{\bibinfo{person}{Junlin Wang}, \bibinfo{person}{Jue Wang}, \bibinfo{person}{Ben Athiwaratkun}, \bibinfo{person}{Ce Zhang}, {and} \bibinfo{person}{James Zou}.} \bibinfo{year}{2024}\natexlab{}.
\newblock \showarticletitle{Mixture-of-Agents Enhances Large Language Model Capabilities}.
\newblock \bibinfo{journal}{\emph{arXiv preprint arXiv:2406.04692}} (\bibinfo{year}{2024}).
\newblock


\bibitem[Wu et~al\mbox{.}(2020)]%
        {wu2020zeroer}
\bibfield{author}{\bibinfo{person}{Renzhi Wu}, \bibinfo{person}{Sanya Chaba}, \bibinfo{person}{Saurabh Sawlani}, \bibinfo{person}{Xu Chu}, {and} \bibinfo{person}{Saravanan Thirumuruganathan}.} \bibinfo{year}{2020}\natexlab{}.
\newblock \showarticletitle{Zeroer: Entity resolution using zero labeled examples}. In \bibinfo{booktitle}{\emph{Proceedings of the 2020 ACM SIGMOD International Conference on Management of Data}}. \bibinfo{pages}{1149--1164}.
\newblock


\bibitem[Xie et~al\mbox{.}(2024)]%
        {xie2024deepmatcher}
\bibfield{author}{\bibinfo{person}{Tao Xie}, \bibinfo{person}{Kun Dai}, \bibinfo{person}{Ke Wang}, \bibinfo{person}{Ruifeng Li}, {and} \bibinfo{person}{Lijun Zhao}.} \bibinfo{year}{2024}\natexlab{}.
\newblock \showarticletitle{Deepmatcher: a deep transformer-based network for robust and accurate local feature matching}.
\newblock \bibinfo{journal}{\emph{Expert Systems with Applications}}  \bibinfo{volume}{237} (\bibinfo{year}{2024}).
\newblock


\bibitem[Xin et~al\mbox{.}(2021)]%
        {xin2021production}
\bibfield{author}{\bibinfo{person}{Doris Xin}, \bibinfo{person}{Hui Miao}, \bibinfo{person}{Aditya Parameswaran}, {and} \bibinfo{person}{Neoklis Polyzotis}.} \bibinfo{year}{2021}\natexlab{}.
\newblock \showarticletitle{Production machine learning pipelines: Empirical analysis and optimization opportunities}. In \bibinfo{booktitle}{\emph{Proceedings of the 2021 International Conference on Management of Data}}. \bibinfo{pages}{2639--2652}.
\newblock


\bibitem[Zhang et~al\mbox{.}(2020)]%
        {zhang2020multi}
\bibfield{author}{\bibinfo{person}{Dongxiang Zhang}, \bibinfo{person}{Yuyang Nie}, \bibinfo{person}{Sai Wu}, \bibinfo{person}{Yanyan Shen}, {and} \bibinfo{person}{Kian-Lee Tan}.} \bibinfo{year}{2020}\natexlab{}.
\newblock \showarticletitle{Multi-context attention for entity matching}. In \bibinfo{booktitle}{\emph{Proceedings of The Web Conference 2020}}. \bibinfo{pages}{2634--2640}.
\newblock


\bibitem[Zhang et~al\mbox{.}(2023)]%
        {zhang2023jellyfish}
\bibfield{author}{\bibinfo{person}{Haochen Zhang}, \bibinfo{person}{Yuyang Dong}, \bibinfo{person}{Chuan Xiao}, {and} \bibinfo{person}{Masafumi Oyamada}.} \bibinfo{year}{2023}\natexlab{}.
\newblock \showarticletitle{Jellyfish: A Large Language Model for Data Preprocessing}.
\newblock \bibinfo{journal}{\emph{arXiv preprint arXiv:2312.01678}} (\bibinfo{year}{2023}).
\newblock


\bibitem[Zhang and Ives(2020)]%
        {zhang2020finding}
\bibfield{author}{\bibinfo{person}{Yi Zhang} {and} \bibinfo{person}{Zachary~G Ives}.} \bibinfo{year}{2020}\natexlab{}.
\newblock \showarticletitle{Finding related tables in data lakes for interactive data science}. In \bibinfo{booktitle}{\emph{Proceedings of the 2020 ACM SIGMOD International Conference on Management of Data}}. \bibinfo{pages}{1951--1966}.
\newblock


\bibitem[Zhang et~al\mbox{.}(2024)]%
        {zhangdirections}
\bibfield{author}{\bibinfo{person}{Zeyu Zhang}, \bibinfo{person}{Paul Groth}, \bibinfo{person}{Iacer Calixto}, {and} \bibinfo{person}{Sebastian Schelter}.} \bibinfo{year}{2024}\natexlab{}.
\newblock \showarticletitle{Directions Towards Efficient and Automated Data Wrangling with Large Language Models}.
\newblock \bibinfo{journal}{\emph{Database and Machine Learning workshop at ICDE}} (\bibinfo{year}{2024}).
\newblock


\end{thebibliography}

\end{document}